\documentclass[twocolumn, switch]{article} 

\usepackage{preprint}

\usepackage{enumitem}
\usepackage{amsmath, amsthm, amssymb, amsfonts}
\usepackage{bm}
\usepackage{amsmath}
\usepackage{graphicx}
\usepackage{subfigure}
\usepackage[algoruled,algo2e]{algorithm2e}

\usepackage{setspace}
\usepackage{amssymb}
\usepackage{booktabs}
\usepackage{array}
\usepackage{color}
\usepackage{multirow}
\usepackage{multicol}
\usepackage{threeparttable}
\usepackage{url}
\usepackage[page]{appendix}

\usepackage[numbers,square]{natbib}
\usepackage[utf8]{inputenc}	
\usepackage[T1]{fontenc}	
\usepackage{xcolor}		
\usepackage[colorlinks = true,
            linkcolor = blue,
            urlcolor  = blue,
            citecolor = blue,
            anchorcolor = black]{hyperref}	
\usepackage{booktabs} 		
\usepackage{nicefrac}		
\usepackage{microtype}		
\usepackage{lineno}		
\usepackage{float}			

\usepackage{newfloat}
\DeclareFloatingEnvironment[name={Supplementary Figure}]{suppfigure}
\usepackage{sidecap}
\sidecaptionvpos{figure}{c}

\usepackage{titlesec}
\titlespacing\section{0pt}{12pt plus 3pt minus 3pt}{1pt plus 1pt minus 1pt}
\titlespacing\subsection{0pt}{10pt plus 3pt minus 3pt}{1pt plus 1pt minus 1pt}
\titlespacing\subsubsection{0pt}{8pt plus 3pt minus 3pt}{1pt plus 1pt minus 1pt}

\usepackage{graphics}
\usepackage{hyperref}
\usepackage{color}
\usepackage{multirow}
\usepackage{hhline}
\usepackage{subfigure}
\usepackage{lipsum}

\title{Improving Generalization of Image Captioning with Unsupervised Prompt Learning}

\usepackage{authblk}

\author[ ]{Hongchen Wei}
\author[*]{Zhenzhong Chen}
\affil[ ]{School of Remote Sensing and Information Engineering, Wuhan University}

\begin{document}

\twocolumn[ 
  \begin{@twocolumnfalse} 
  
\maketitle

\begin{abstract}

Pretrained visual-language models have demonstrated impressive zero-shot abilities in image captioning, when accompanied by hand-crafted prompts. 
Meanwhile, hand-crafted prompts utilize human prior knowledge to guide the model. 
However, due to the diversity between different domains, such hand-crafted prompt that provide invariant prior knowledge may result in mode collapse for some domains. 
Some researches attempted to incorporate expert knowledge and instruction datasets, but the results were costly and led to hallucinations. 
In this paper, we propose an unsupervised prompt learning method to improve Generalization of Image Captioning (GeneIC), which learns a domain-specific prompt vector for the target domain without requiring annotated data. 
GeneIC aligns visual and language modalities with a pre-trained Contrastive Language-Image Pre-Training (CLIP) model, thus optimizing the domain-specific prompt vector from two aspects: attribute and semantic consistency. 
Specifically, GeneIC first generates attribute-transferred images with differing attributes, while retaining semantic similarity with original images. 
Then, GeneIC uses CLIP to measure the similarity between the images and the generated sentences. 
By exploring the variable and invariant features in the original images and attribute-transferred images, attribute consistency constrains the attribute change direction of both images and sentences to learn domain-specific knowledge. 
The semantic consistency directly measures the similarity between the generated sentences and images to ensure the accuracy and comprehensiveness of the generated sentences. 
Consequently, GeneIC only optimizes the prompt vectors, which effectively retains the knowledge in the large model and introduces domain-specific knowledge. 
Experiments show that GeneIC exhibits superior generalization performance compared to state-of-the-art methods on multiple target domain datasets.

\end{abstract}

\vspace{0.4cm}

  \end{@twocolumnfalse} 
] 

\newcommand\blfootnote[1]{%
\begingroup
\renewcommand\thefootnote{}\footnote{#1}%
\addtocounter{footnote}{-1}%
\endgroup
}

\section{INTRODUCTION}

{\blfootnote{Corresponding author: Zhenzhong Chen, E-mail:zzchen@ieee.org}}
  
  Recent advances in pretrained Visual-Language Models (VLMs)~\citep{JiaYXCPPLSLD21, RadfordKHRGASAM21, 0001LXH22, AlayracDLMBHLMM22, li2023blip2} have undergone significant advancements and have achieved remarkable performance on different downstream tasks, such as image captioning~\citep{VinyalsTBE15,YangZGC23,WuZL22}, which aims to automatically generate captions for images. 
  Meanwhile, some of these models~\citep{AlayracDLMBHLMM22, li2023blip2, abs-2305-06500, abs-2304-10592} have demonstrated impressive zero-shot capability in image captioning, requiring only hand-crafted prompts. 
  This is highly beneficial, as it eliminate the model's dependence on downstream target domain image-text pair data, thereby, providing a viable research solution for domain generalization image captioning, which is training on a source domain and generalizing to any domain. 
  
\begin{figure}[!t]
      \centering
      \includegraphics[width=3.2in]{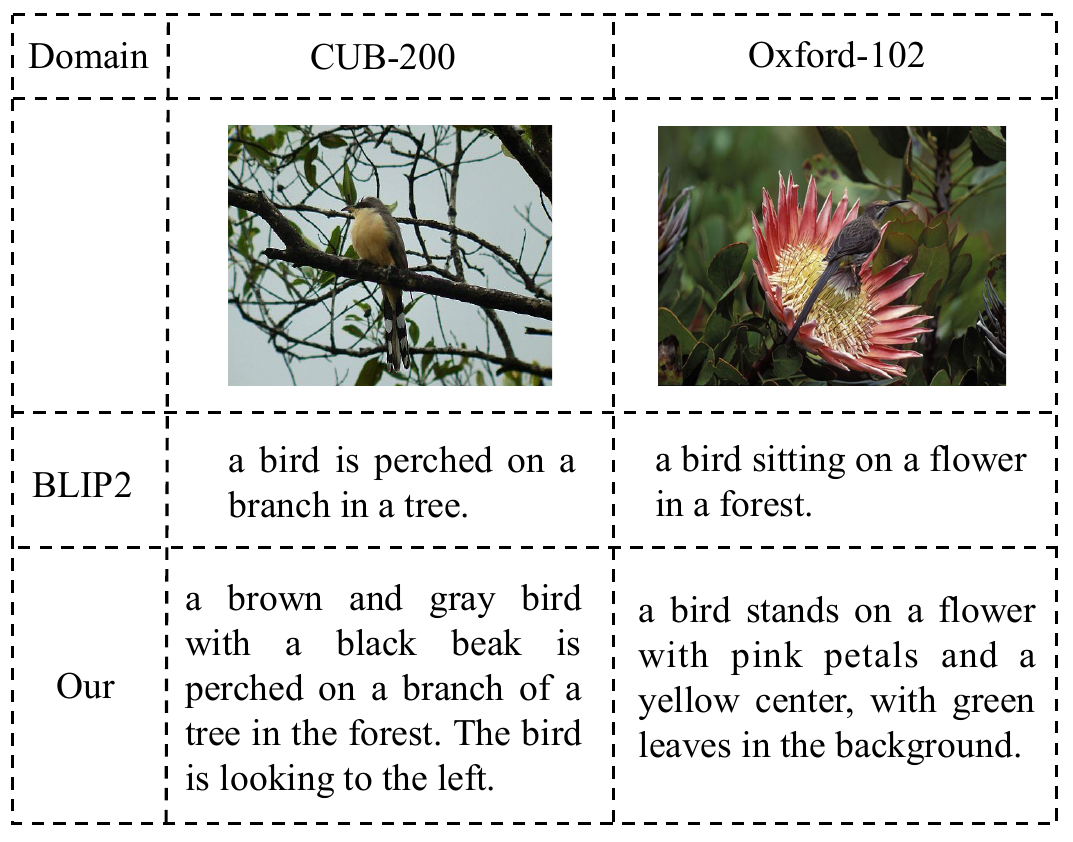}
      \caption{Example of mode collapse. 
      Previous methods, such as BLIP2, use hand-crafted prompt ($e.g.$, ``A photo of'') to generate sentences for image from different target domains, such as CUB-200 (bird data)~\citep{ReedALS16} and Oxford-102 (flower data)~\citep{NilsbackZ08}. 
      The generated sentences exhibit similar modes, and lacking descriptions of domain-specific knowledge such as bird body features and flower color attributes, resulting in limited diversity and low quality. 
      In contrast, our approach learns domain-specific prompt vectors for different target domains by exploring both variable and invariant features present in the target domain. 
      This guides the model to generate sentences that incorporate domain-specific knowledge.}
      \label{fig_1}
  \end{figure}

  However, recent research~\citep{ZhouYLL22,ZhouYL022,abs-2210-03117,ShuNHYGAX22,Jin0SC022,abs-2304-03119} indicates that hand-crafted prompts might be suboptimal. 
  On the one hand, the model is sensitive to prompt, and slight variations in wording could make a large difference in performance. 
  Meanwhile, effective prompt crafting requires a comprehension of the previous knowledge associated with downstream tasks and the underlying mechanism of the model. 
  On the other hand, when dealing with data from different domains, such prompt guide the model with the same prior knowledge, making it unable to adaptively distinguish the differences between different domains and thus overlooking the domain-specific knowledge of the target domain. 
  This may lead to mode collapse. 
  As shown in Figure \ref{fig_1}, traditional methods utilized hand-crafted prompt. 
  However, hand-crafted prompts failed to take into account the distinct domain-specific knowledge present within each target domain dataset, such as bird wing and beak features or flower colors, which consequently make generated descriptions that lack diversity and specialized knowledge.
  This hinders the development of domain generalization for image captioning. 
  
  One of the tasks associated with domain generalization for image captioning is cross-domain image captioning~\citep{ChenLCHFS17,0007ZXFZ0L19,YuanZHZXLW22}, which is trained on a source domain and  a small amount of target domain data, and generalized to target domain. 
  In details,~\citep{ChenLCHFS17} explored a discriminator network in adversarial learning to evaluate the similarity between generated captions and target domain captions. 
 ~\citep{0007ZXFZ0L19} was first pre-train model, and then fine-tuned with a small amount of target domain dataset. 
 ~\citep{YuanZHZXLW22} directly integrated target domain information into the model. 
  However, it is important to note that these methods only generalize to one specific target domain, which makes them less applicable than domain generalization for image captioning. 
  Moreover, existing cross-domain methods often require introducing additional target domain image-caption priors to align the caption styles between the source and target domains. 
  For some scenarios where data collection is challenging, such as art and medicine, the viability of these methods may be negative. 
  
  Inspired by prompt learning research in Natural Language Processing (NLP)~\citep{ShinRLWS20,ZhouHMBN22,JoulinMJV16}, some studies~\citep{ZhouYLL22,abs-2210-03117,ShuNHYGAX22,ZhouYL022,abs-2304-03119} attempted to automate prompt engineering in pretrained visual-language models. 
  Specifically, they modeled the context of prompt with learnable vectors.
  For example,~\citep{ZhouYLL22} used only minimal amount of target domain data to learn a specific set of context tokens for each domain. 
 ~\citep{abs-2210-03117} designed learnable prompts for both image and text modalities.  
  While the methods mentioned above have yielded favorable results, they still require a small amount of target domain data optimization the prompt vector. 
  In addition, they have primarily centered around classification tasks, without taking into account the complexity and dependence on context involved in sequence generation tasks. 
  To address this issue, other studies have attempted to incorporate instruction tuning~\citep{WeiBZGYLDDL22,abs-2210-11416} in generative visual-language models. 
  These methods fine-tune the model to enhance the recognition of hand-crafted prompts. 
  ~\citep{abs-2304-10592} collected a high-quality and well-aligned dataset of conversations. 
  ~\citep{abs-2305-06500} converted public datasets into an instruction-caption format for fine-tuning the model. 
  However, it resulted in hallucination~\citep{abs-2305-10355} and the resource consumption, a more meaningful direction would be to combine unsupervised learning for achieving domain generalization for image captioning.
  
  Motivated by this, this paper presents an unsupervised prompt learning method to improve Generalization of Image Captioning (GeneIC), which learns a domain-specific prompt vector for the target domain without requiring annotated data. 
  Unlike single-modal tasks, the challenge of cross-modal unsupervised learning lies in the heterogeneity between different modalities. 
  To address this issue, GeneIC utilizes a pre-trained CLIP model~\citep{RadfordKHRGASAM21} to project both visual and language content into a shared semantic space. 
  Domain-specific prompt vectors are then optimized from two perspectives: attribute and semantic consistency. 
  The former learns domain-specific knowledge, while the latter aligns visual and language content. 
  Specifically, for attribute consistency, GeneIC first generates attribute-transferred images, which are obtained by modifying the feature maps of the original target domain images in the autoencoder. 
  These images have similar semantics as the original, but with different attributes, like a bird with black wings and a bird with blue wings. 
  By exploring the variable and invariant features in the original images and attribute-transferred images, attribute consistency constrains the attribute change direction of both images and sentences in the CLIP space to learn domain-specific knowledge. 
  For semantic consistency, GeneIC directly minimizes the distance between input images and generated sentences in the CLIP space to ensure semantic consistency and produce accurate sentences.
  Furthermore, to retain the knowledge in the pre-trained model, GeneIC freezes the majority of the parameters in the model and optimizes only the prompt vectors. 
  Compared with methods such as~\citep{abs-2305-06500,abs-2304-10592}, our approach requires fewer training resources. 
  
  In summary, the contributions of this paper can be summarized as follows:
  \begin{enumerate}
    \item We explored an unsupervised prompt learning method to improve generalization of image captioning, which learns a domain-specific prompt vector for target domain without requiring annotated data. 
      The prompt vector is employed to guide the model to generate captions that incorporate domain-specific knowledge, thus alleviates mode collapse. 
    \item We designed novel attribute and semantic consistency to optimize the prompt vector. 
    The former explore the variable and invariant features in the target domain, to constrains the attribute change direction of both images and sentences to learn domain-specific knowledge. 
    The latter directly enforces the semantic coherence between input images and generated sentences, thereby enhancing the accuracy of generated sentences. 
    \item Based on prompt learning, our method utilizes domain-specific prompt vectors instead of hand-crafted prompts, achieving superior generalization performance. 
    Additionally, our method is more parameter-efficient than traditional methods because it optimized a small number of parameters. 
  \end{enumerate}
  
  The remainder of this paper is organized as follows. 
  In Section~\ref{sec:related_work}, we provide an overview of related work. 
  In Section~\ref{sec:method}, each module and loss item in GeneIC is introduced in details. 
  In Section~\ref{sec:experiments}, we present the experimental setup of GeneIC, and the experimental results are analyzed. 
  Finally, Section~\ref{sec:conclusion} provides a brief summary of the paper. 
  
  \section{Related Work} \label{sec:related_work}
  
  \subsection{Domain Generalization for Image Captioning} \label{sec:related_work;subsec:vl_models}
  Domain generalization for image captioning~\citep{ren2023crossing} aims to generate descriptions for target domain images in scenarios where target domain annotations are not available. 
  The major challenge is seizing specific features in data of target domains. 
  A related task is cross-domain image captioning, which introduces the target domain into the training process to alleviate domain shift. 
  For instance,~\citep{ChenLCHFS17} utilizes adversarial networks to discriminate whether the generated captions correspond to the target domain. 
 ~\citep{0007ZXFZ0L19} designed a multi-task learning strategy to optimize both image captioning and image synthesis. 
 ~\citep{ZhaoWL21} proposed a cross-domain image captioning method based on a retrieval model to promote domain adaptation of the model. 
 ~\citep{YuanZHZXLW22} proposed a style-based cross-domain image captioning method that integrates style information into the model. 
  Compared with domain generalization for image captioning, the task have the following drawbacks. 
  First, to align the source domain and target domain, additional target domain caption priors or even a small quantity of target domain image-text pairs must be introduced. 
  However, these methods are not applicable in some scenarios, such as art and medicine, where gathering data is challenging.  
  Secondly, the model inference process is only capable of generalizing to a single target domain, meaning that when a new target domain arises, the model must be retrained, which can be very resource-intensive. 
  
  With the emergence of large-scale pre-training models, researchers have proposed zero-shot image captioning. 
 ~\citep{TewelSSW22} merged the CLIP and language model to complete zero-shot image captioning in any domain without the need for training the model. 
 ~\citep{abs-2303-02437} aligned visual content through induction of CLIP-induced score affecting the language model creation. 
 ~\citep{abs-2211-07275} proposed anchor enhancement to guide the generation model to focus on fine-grained information in the representation of CLIP. 
  Despite the progress made, the performance of image captioning in specific domains is still low and lacks descriptions of domain-specific knowledges. 
  One of the important reasons for this issue is the large semantic gap between the CLIP model and language model. 
  
  To address this issue, several studies~\citep{AlayracDLMBHLMM22,abs-2302-14045,abs-2303-16199,li2023blip2} have introduced a joint training method of visual and language models utilizing large-scale image-text paired datasets. 
  These models have demonstrated impressive zero-shot abilities in image captioning, when accompanied by hand-crafted prompts. 
  For example,~\citep{AlayracDLMBHLMM22} proposed a new cross-attention layer and inserted it into pre-trained large language model to train the model on billions of image-text pairs. 
 ~\citep{abs-2302-14045} designed a large multimodal language model that can process any interleaved text and images. 
 ~\citep{abs-2303-16199} proposed a zero-init attention mechanism with zero-gating, which can adaptively inject new instructional clues into the language model. 
 ~\citep{li2023blip2} used a lightweight query transformer to bridge the visual and language models.
  However, recent research~\citep{ZhouYLL22,ZhouYL022,abs-2210-03117,ShuNHYGAX22,Jin0SC022,abs-2304-03119} presents that hand-crafted prompts may suboptimal. 
  On the one hand, the words highly influence hand-crafted prompts, and slight variations in wording may have a significant effect on results. 
  On the other hand, hand-crafted prompt, such as ``A photo of'', result in strict constraints for different target domain datasets, make models to generate homogeneous and low-quality captions ($i.e.$, ``mode collapse''). 
  To improve the model's prompt robustness, ~\citep{abs-2304-10592,abs-2305-06500} developed instruction datasets for fine-tuning the model. 
  However, this approach inevitably entails substantial resource consumption. 
  Additionally, research~\citep{abs-2305-10355} has revealed that instruction datasets may induce hallucinations.

  \subsection{Prompt Learning for Vision-Language Models} \label{sec:related_work;subsec:prompt_nlp}
  In recent years, Large Language Models (LLM)~\citep{BrownMRSKDNSSAA20,abs-2205-01068,abs-2302-13971,WeiBZGYLDDL22} have exhibited remarkable capabilities in generating language. 
  Efficiently fine-tuning the LLM model has become a hot research topic in order to better generalize to downstream tasks. 
  Among them, prompt learning only trains additional prompt vectors under the condition of frozen model parameters, which can guide the model to better adapt to downstream tasks. 
  
  Inspired by prompt learning in Natural Language Processing (NLP)~\citep{HoulsbyGJMLGAG19,RaffelSRLNMZLL20,LiL20}, some studies~\citep{ZhouYLL22,abs-2210-03117,ShuNHYGAX22,abs-2204-03649,abs-2210-07225} have attempted to introduce prompt learning into multimodal visual language models to replace hand-crafted prompts and improve classification performance on target domain data. 
  Specifically,~\citep{ZhouYLL22} proposed a context-aware optimization method that uses learnable vectors to model contextual words in prompts. 
 ~\citep{abs-2210-03117} designed a multimodal prompt learning method, with visual and text encoders learning prompt vectors separately. 
 ~\citep{ShuNHYGAX22} used a single-sample dynamic learning adaptive prompt in the testing process. 
 ~\citep{abs-2204-03649} proposed an unsupervised prompt learning method to improve the transfer performance of the CLIP model. 
 ~\citep{abs-2210-07225} combines the advantages of textual and visual prompts and proposes a unified prompt tuning method. 
  The above methods have achieved good performance in domain generalization for classification tasks. 
  However, compared with classification tasks, image captioning~\citep{LiCL19,li2018image,abs-2003-10925} needs to consider the complexity of text sequences and contextual dependencies, and is therefore more challenging. 
  
  \section{Methodology} \label{sec:method}
  
  In this section, we propose an unsupervised prompt learning to improve generalization of image captioning, which learns a domain-specific prompt vector for the target domain without requiring annotated data. 
  Compared with traditional methods, our method has several advantages. 
  First, we can generalize the model to any target domain without the need for annotated data. 
  Second, we only optimize the prompt vectors, which can achieve superior generalization performance at a lower computational cost. 

  Next,  we will provide a detailed introduction to this method. 
  Firstly, we introduce the data composition of the domain generalization for image captioning and the backbone model. 
  Afterwards, we elaborate on the proposed framework in details, including attribute consistency and semantic consistency. 
  
  \begin{figure*}[t]
    \centering
    \includegraphics[width=155mm]{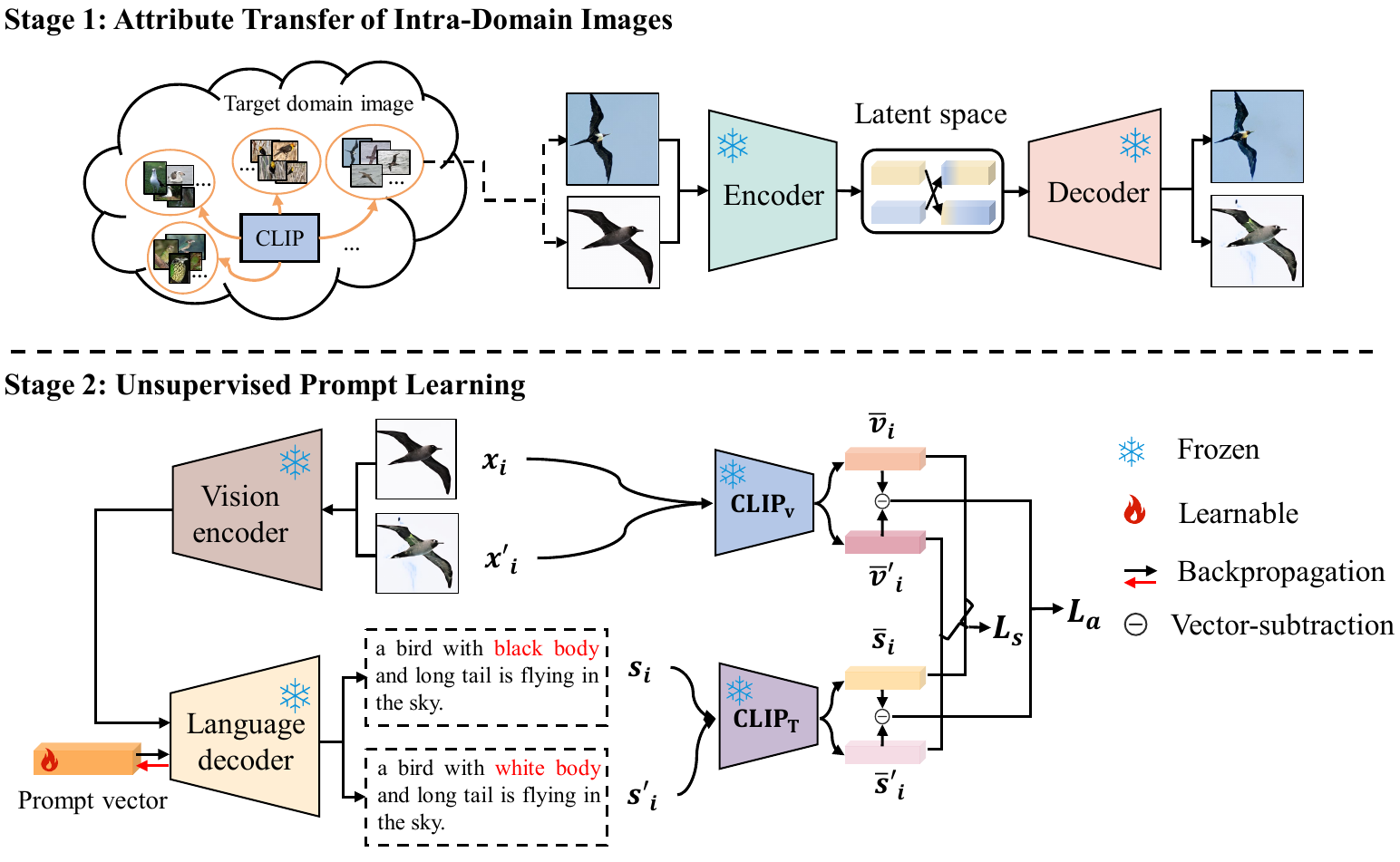}
    \caption{The illustration of GeneIC. 
    In the first stage, the pre-trained CLIP model is used to cluster images in the target domain. 
    Then, two similar images are input into the pre-trained VQ-GAN model to extract feature maps, and their partial feature maps are exchanged to achieve meaningful attribute transfer. 
    Finally, attribute-transferred images are reconstructed. 
    In the second stage, the original image and the attribute-transferred image are input as an image pair into a pre-trained visual language model ($e.g.$, BLIP2) with frozen parameters to generate corresponding sentences. 
    The images and sentences are projected into the same space using CLIP, and the prompt vector is optimized through attribute consistency ($L_a$) and semantic consistency ($L_s$). 
    }\label{fig_2}
  \end{figure*}
  
  \subsection{Overview of GeneIC}
  In domain generalization for image captioning, only the target domain image set $X = {\left \{ \boldsymbol{x}_{i} \right \}}^{N}_{i=1}$ is used, without any labeled data, where $\boldsymbol{x}_{i}$ represents the $i$-th image. 
  Without any loss of generality, GeneIC can take any state-of-the-art pre-trained visual-language models. 
  Considering the effectiveness, in this paper, GeneIC takes BLIP2~\citep{li2023blip2} as backbone. 
  BLIP2 is a powerful visual-language model that is at the forefront of the field of zero-shot learning. 
  It has demonstrated remarkable proficiency in overcoming challenges related to limited data availability through its prompt-based ability to adapt rapidly to new tasks or concepts.
  Next, we provide a concise overview of the training and inference processes employed by BLIP2. 
  
  BLIP2 employs the pre-trained visual encoder and Large Language Model (LLM), and trains a lightweight query transformer to connect different modalities, which demonstrates remarkable generality and efficiency. 
  The visual encoder is based on the CLIP model, which is trained on billions of image-text pairs using contrastive loss and is highly effective in aligning vision and language. 
  In addition, the language model employs the unsupervised-trained OPT~\citep{abs-2205-01068} model family and the instruction-trained FlanT5~\citep{abs-2210-11416} model family. 
  This paper adopts pre-trained OPT2.7B as the language model. 
  Specifically, BLIP2 includes two-stage training. 
  In the first stage, BLIP2 combines the query transformer (QFormer) with the frozen visual encoder to learn representation through the proxy task, to retain critical visual information. 
  In the second stage, the query transformer is attached to the frozen LLM to generate text, and its output is linearly projected to the embedding space of the LLM through a fully-connected layer.
  These visual cues guide the generation process and enable learning of visual-language alignment through generation loss.
  During the inference process, with image captioning as an example, the test image ($i.e.$, $\boldsymbol{x}_i$) is initially inputted into the visual encoder and query transformer to produce the visual embedding, $\boldsymbol{v}_i$. 
  To ensure that the generated captions accurately match the user's intention, BLIP2 utilizes hand-crafted prompts to guide the model. 
  More specifically, the hand-crafted prompt is projected into the embedding space of LLM. Together with $\boldsymbol{v}_i$, they are then inputted into LLM for generating the captions.
  The formula is expressed as follows: 
  \begin{equation} \label{eq1}
    \begin{split}
      \boldsymbol{v}_i & = \operatorname{encoder}(\boldsymbol{x}_i) \\
      \boldsymbol{\hat{v}}_i & = \operatorname{concat}(W_v*\boldsymbol{v}_i, W_p*\boldsymbol{p}) \\
      \boldsymbol{s}_i & = \operatorname{decoder}(\boldsymbol{\hat{v}}_i)
    \end{split}
  \end{equation}
  where the encoder comprises a pre-trained visual decoder and a query transformer, with $W_v$ symbolizing a fully connected layer to facilitate mapping visual information to the LLM space. 
  $\boldsymbol{p}$ denotes hand-crafted prompt ($e.g.$, ``A photo of'') and $W_p$ maps $\boldsymbol{p}$ to the LLM space. The concatenated vector is denoted as $\boldsymbol{\hat{v}}_i$, which is then fed into the LLM decoder to generate the sentence $\boldsymbol{s}_i$.
  
  However, recent research~\citep{ZhouYLL22,ZhouYL022,abs-2210-03117,ShuNHYGAX22,Jin0SC022,abs-2304-03119} indicates that models are very sensitive to hand-crafted prompts. 
  Wording can noticeably affect the performance. 
  In addition, using hand-crafted prompts to generate captions makes limited diversity and low quality. 
  The reason is that hand-crafted prompts cannot adaptively guide the model to focus on domain-specific knowledge when dealing with data from different domains, due to their invariant prior knowledge. 
  Taking inspiration from prompt learning~\citep{ZhouYLL22,ZhouYL022,abs-2210-03117,ShuNHYGAX22}, a possible solution would be to learn prompt vectors instead of to design hand-crafted prompt. 
  As a result, we propose an unsupervised prompt learning method (GeneIC), to enhance the generalization ability of image captioning models. 
  
  Figure \ref{fig_2} is the pipeline of GeneIC, which includes two stages. 
  In the first stage, GeneIC achieves domain-specific attribute transfer, generating images that are semantically similar to the original images in the domain, but with different attributes. 
  Specifically, GeneIC uses a pre-trained Vector Quantised Generative Adversarial Network (VQ-GAN)~\citep{EsserRO21} to project images into latent space, and achieves attribute transfer between target domain images by swapping partial feature maps between the same mini-batch samples. 
  To ensure meaningful attribute transfer, GeneIC first clusters the test images of the target domain using CLIP, with similar samples taken as a mini-batch. 
  Then, it explores the semantics of each feature map and replaces the feature map of the main objectives ($e.g.$, bird, instead of background). 
  Finally, the attribute-transferred image is decoded. 
  In the second stage, a pair of images consisting of original images and attribute-transferred images are inputted into a parameter-frozen visual-language model. 
  To better generalize the model to the target domain, a learnable prompt vector is used instead of hand-crafted prompt to guide the model to better focus on the knowledge of the target domain. 
  GeneIC uses CLIP to project the input image and the generated sentence into the same semantic space, optimizing the prompt vector with attribute consistency ($L_a$) and semantic consistency ($L_s$). 
  The former aims to explore domain-specific attributes, while the latter focuses on the all elements in the image. 
  Next, we will provide a detailed introduction of the two stages. 
  
  \begin{figure}[t]
    \centering
    \includegraphics[width=80mm]{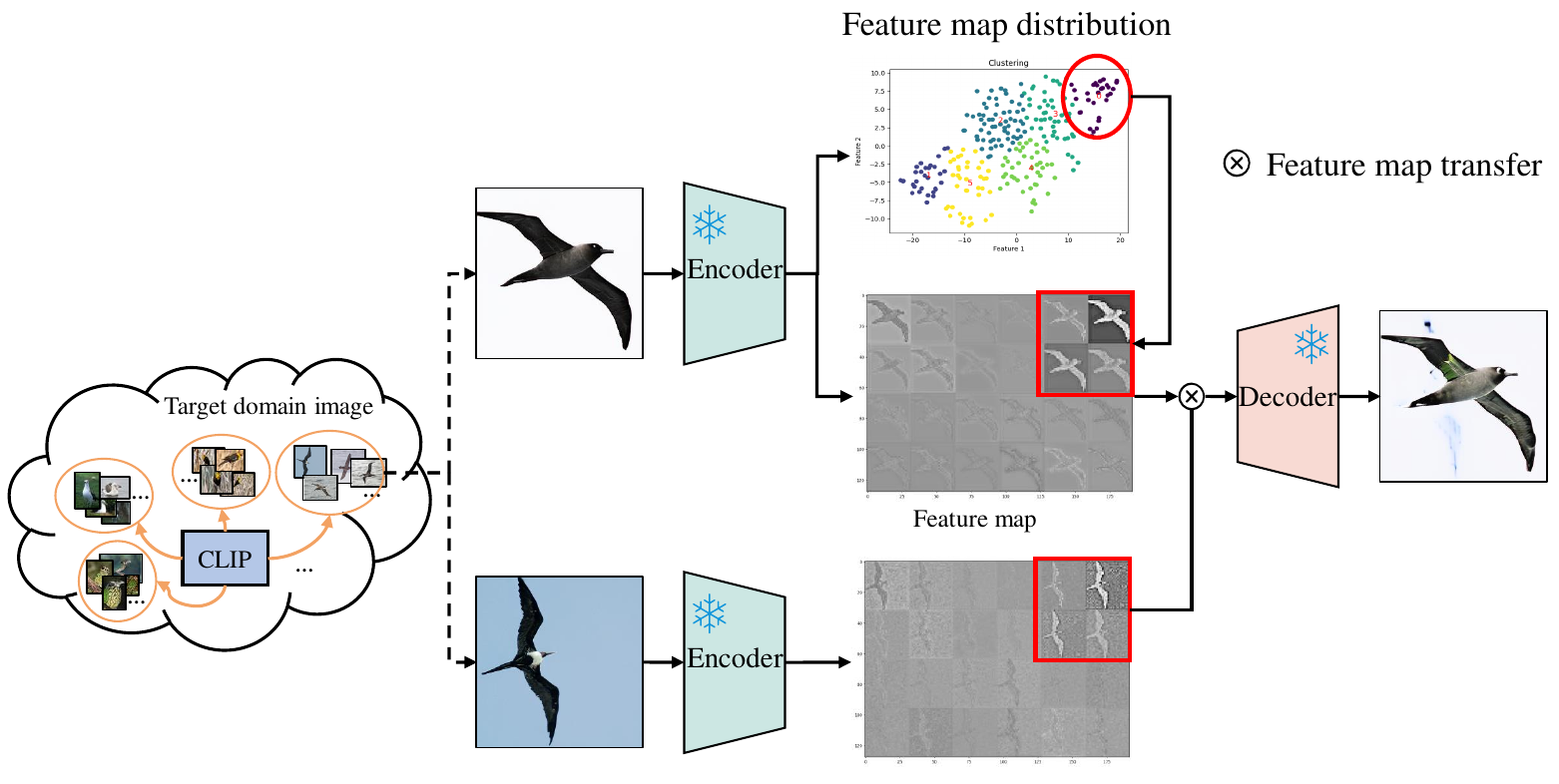}
    \caption{The pipeline of attribute transfer of intra-domain images. 
    We begin by utilizing CLIP model to cluster images in the target domain. 
    This is to avoid introducing additional background noise. 
    Then, we input two semantically similar images in an encoder, resulting in two feature map sets. 
    Next, we replace the feature map that contains the main objectives knowledge ($i.e.$, the red box, which is retrieved from the feature map distribution and corresponds to a red circle), to achieve meaningful attribute modification. 
    Finally, we input the modified feature map into the decoder, to generate an image that incorporates the modified features.}
    \label{fig_3}
  \end{figure}
  
  \subsection{Attribute Transfer of Intra-Domain Images}
  While hand-crafted prompts ($e.g.$, ``A photo of") exhibit well performance on source domain~\citep{li2023blip2,AlayracDLMBHLMM22,abs-2302-14045}, they fail to adapt efficiently to target domains with significant domain shift from the source domain~\citep{ZhouYLL22,abs-2210-03117,ShuNHYGAX22}. 
  Hand-crafted prompts lack the ability to distinguish between the source and target domains. 
  Thus, they often overlook fine-grained attributes in the target domain and produce rough captions to describe the images. 
  
  To capture domain-specific attributes more accurately, it is intuitive to learn attribute differences among target domain images. 
  In this paper, we propose an unsupervised image attribute transfer method that employs feature map swapping to produce images with semantic similarities but different attributes for the target domain images. 
  By examining the attribute modifications between the original images and the attribute-transferred images, the model can learn domain-specific attributes. 
  In order to obtain meaningful attribute transfer, we explore the feature maps in the latent space. 
  Previous studies~\citep{ZhangWZ18a,BauZKO017} suggest that information represented by diverse feature maps in convolutional neural networks is distinct, for example, color, texture, and edges. 
  Therefore, we concentrate on the feature maps of the main objectives ($e.g.$, birds in CUB-200 dataset) in the target domain images, and obtain meaningful attribute transfer by modifying the feature maps. 
  To ensure the quality of reconstructed images, we use the pre-trained VQ-GAN~\citep{EsserRO21} as the backbone. 

  As shown in Figure \ref{fig_3}, we begin by clustering the target domain images using the CLIP model. 
  This is to avoid introducing additional background noise. 
  Then, we input two similar images into the encoder as a batch to obtain the feature map sets $\boldsymbol{f_i}$ and $\boldsymbol{f_j}$. 
  We exchange attributes of semantically similar images only, which helps avoid producing meaningless noise attributes resulting from large semantic differences.
  \begin{equation} \label{eq2}
    \begin{split}
      \{\boldsymbol{f}_i, \boldsymbol{f}_j\}=&\operatorname{encoder}\left(\boldsymbol{x}_i, \boldsymbol{x}_j\right), \{\boldsymbol{f}_i, \boldsymbol{f}_j\} \in \mathbb{R} ^{l*w*c}
    \end{split}
  \end{equation}
  where $\boldsymbol{x}_i$ and $\boldsymbol{x}_j$ represents similar images from the target domain, $\boldsymbol{f_i}$ and $\boldsymbol{f_j}$ represents a set of feature maps, and $l$, $w$ and $c$ represents the length, width and channels of the feature maps. 
  
  Based on~\citep{ZhangWZ18a,BauZKO017}, we extract the feature map of the main objectives, replacing it with the feature map of a similar image to accomplish meaningful attribute modification. 
  Figure \ref{fig_3} illustrates the feature map transfer operation, which showcases this process. 
  By tsne~\citep{vanDerMaaten2008} visualizing the distribution of feature maps in the target image, we observed that the main objectives feature maps are concentrated in the upper right corner of the distribution ($i.e.$, the red circle). 
  The visualization of the feature maps verifies this conclusion, as the feature maps represented by the red circle show more attention to the main objectives compared to other regions in the distribution. 
  Through modifying these feature maps, we can attain attribute modifications of the main objectives in the image. 
  Therefore, we use feature maps from similar images in the same positions to substitute these feature maps and achieve attribute modifications. 
  The feature map exchange process is as follows:
  \begin{equation} \label{eq3}
    \begin{split}
      \{\boldsymbol{f}^r_i, \boldsymbol{f}^r_j\}=&\operatorname{retrieval}\left(\boldsymbol{f}_j, \boldsymbol{f}_i\right), \{\boldsymbol{f}^r_i,  \boldsymbol{f}^r_j\} \in \mathbb{R} ^{l*w*c_r} \\
      \boldsymbol{f}_{i}^{\prime}=&\operatorname{transfer}\left(\boldsymbol{f}_i, \boldsymbol{f}^r_i, \boldsymbol{f}^r_j \right), \boldsymbol{f}_i^{\prime} \in \mathbb{R} ^{l*w*c} 
  \end{split}
  \end{equation}
  where the $\operatorname{retrieval}$ operation involves retrieval for feature maps that contain information about the main objectives, specifically in the top-right corner of their distribution. 
  $c_r \in  c$ denotes the number of feature maps that contain information about the main objectives. 
  The $\operatorname{transfer}$ operation involves replacing the original set of feature maps, $\boldsymbol{f_i}$, with $\boldsymbol{f}^r_i$, using the feature maps $\boldsymbol{f}^r_j$ from similar images at the same position. 
  
  Finally, we input the new feature map set into the decoder. 
  \begin{equation} \label{eq4}
    \begin{split}
      \boldsymbol{x}_{i}^{\prime}=&\operatorname{decoder}\left({\boldsymbol{f}}_i^{\prime} \right)
    \end{split}
  \end{equation}
  where $\boldsymbol{f_i}^{\prime}$ represents the set of feature maps used for attribute transfer, while $\boldsymbol{x}_i^{\prime}$ represents the generated attribute-transferred image. 
  
  \subsection{Unsupervised Prompt Learning}
  To enhance generalization to downstream tasks, downstream data is typically used to fine-tune pre-trained models and align the source domain distribution with the target domain distribution. 
  Two issues arise in this method. 
  The first issue pertains to constructing a supervised downstream dataset. 
  The construction of paired data comprising images and text necessitates significant resources. 
  The second concern is fine-tuning large-scale models. 
  Previous research suggests~\citep{LiL20,LesterAC21} that fine-tuning large-model parameters on small-scale downstream data can result in catastrophic forgetting. 
  
  Therefore, we propose an unsupervised prompt learning method to improve generalization of image captioning. 
  It freezes all the parameters of a pretrained model and only utilizes unlabeled target domain images to optimize prompt vectors. 
  As shown in the second stage of Figure \ref{fig_2}, where two loss functions are employed to optimize the prompt vector: attribute consistency $L_a$ and semantic consistency $L_s$. 
  The former learns domain-specific attribute knowledge, while the latter ensures the accuracy and comprehensive of generated sentences. 
  Subsequently, we will provide a detailed explanation of prompt learning and two kinds of unsuperivsed loss: attribute consistency $L_a$ and semantic consistency $L_s$. 
  
  \subsubsection{Prompt Learning}
  Traditional large visual-language models typically depend on hand-crafted prompts to differentiate downstream tasks. 
  For example, BLIP2 utilizes the phrase ``A photo of" as hand-crafted prompt in image captioning, showcasing exemplary zero-shot performance on the MSCOCO~\citep{LinMBHPRDZ14} and Flickr30k~\citep{YoungLHH14} datasets. 
  However, hand-crafted prompts underperform in target domain datasets that exhibit significant differences from source domain, as they are incapable of adaptively focusing on fine-grained attributes within the target domain. 
  
  Inspired by~\citep{ZhouYLL22,ZhouYL022,abs-2210-03117}, we construct learnable prompt vectors that adaptively learn domain-specific knowledge. 
  The pre-trained BLIP2 serves as our backbone model. 
  While training, we freeze all model parameters and only optimize learnable prompt vectors. 
  More specifically, $\boldsymbol{p}_v$, a prompt vector randomly initialized, is concatenated with visual embedding $\boldsymbol{v}_i$ and inputted together into the language decoder to generate corresponding captions, $\boldsymbol{s}_i$. 
  The formula is expressed as:
  \begin{equation} \label{eq5}
    \begin{split}
      \boldsymbol{v}_i & = \operatorname{encoder}(\boldsymbol{x}_i) \\
      \boldsymbol{\hat{v}}_i & = \operatorname{concat}(W_v*\boldsymbol{v}_i, \boldsymbol{p}_v) \\
      \boldsymbol{s}_i & = \operatorname{decoder}(\boldsymbol{\hat{v}}_i)
    \end{split}
  \end{equation}
  
  To learn domain-specific knowledge, we introduce two kinds of unsuperivsed loss to optimize the prompt vector: attribute consistency and semantic consistency. 
  
  \subsubsection{Attribute Consistency}
  Examining the disparities between target domain images, particularly those that exhibit semantic similarities but vary in attributes, is an intuitive approach for acquiring domain-specific knowledge. 
  For instance, within the CUB-200 dataset, birds manifest distinct colors, necessitating a focus on the alterations in their attributes rather than those in the background. 
  When an image's attributes undergo modification, the generated sentences should align with these alterations. 
  Utilizing this concept, we devised an unsupervised attribute consistency loss. 
  Specifically, we employ the original image ($i.e.$, $\boldsymbol{x}_i$) and the attribute-transferred image ($i.e.$, $\boldsymbol{x}_i^{\prime}$) from the target domain, as input image pairs for the model. 
  To precisely calculate attribute changes and ensure the consistency of heterogeneous modes, we extract the features of input image pairs and generated sentences with pre-trained CLIP model. 
  By projecting images and captions into the same semantic space with CLIP, we are able to directly compare the two heterogeneous modalities. 
  Attribute transfer consistency is defined with the following equation: 
  \begin{equation} \label{eq6}
    \begin{split}
      \Delta \mathrm{V}_i =& \operatorname{Norm}\left(\boldsymbol{\bar{v}}_i\right)-\operatorname{Norm}\left(\boldsymbol{\bar{v}}_{i}^{\prime}\right) \\
      \Delta \mathrm{S}_i =& \operatorname{Norm}\left(\boldsymbol{\bar{s}}_{i}\right)-\operatorname{Norm}\left(\boldsymbol{\bar{s}}_{i}^{\prime}\right) 
    \end{split}
  \end{equation}
  where $\boldsymbol{\bar{v}}_i, \boldsymbol{\bar{v}}_i^{\prime} = \operatorname{CLIP_V}\left(\boldsymbol{x}_i, \boldsymbol{x}_i^{\prime} \right)$ represent the original image and attribute transfer image features extracted using CLIP, and $\boldsymbol{\bar{s}}_i, \boldsymbol{\bar{s}}_i^{\prime} = \operatorname{CLIP_T}\left(\boldsymbol{s}_i, \boldsymbol{s}_i^{\prime} \right)$  represent the corresponding caption features. 
  $\Delta \mathrm{V}_i$ and $\Delta \mathrm{S}_i$ represent the attribute changes between images and captions, respectively. 
  $\operatorname{Norm}\left(\cdot \right)$ represents $L_2$ normalization. 
  We use the $L_a$ constraint to ensure consistency between the attribute changes of images and sentences,  and by explore variable and invariant features to learn domain-specific knowledge.
  \begin{equation} \label{eq7}
    L_a = \mathbb{E} \sum_{i=1}^n\left(1 - \frac{\Delta \mathrm{V}_i \cdot \Delta \mathrm{S}_i}{\left|\Delta \mathrm{V}_i \right|\left|\Delta \mathrm{S}_i\right|}\right)
  \end{equation}
  where $n$ is the batch size. 
  
  \subsubsection{Semantic Consistency}
  Recent research~\citep{00010KDBB22} demonstrates that matching scores the image and text based on CLIP can increase the diversity and accuracy of captions generated. 
  This enhancement is achieved through direct measurement of the correlation between the input image and the generated sentences. 
  Thus, we propose a semantic consistency loss. 
  By leveraging knowledge in an open world, we aim to enhance the quality and comprehensive of generated captions. 
  The formula for the semantic consistency loss term is expressed as follows: 
  \begin{equation} \label{eq8}
    \begin{split}
      L_s=\mathbb{E} \sum_{i=1}^n\left(1 - \frac{\boldsymbol{\bar{v}}_i \cdot \boldsymbol{\bar{s}}_{i}}{\left|\boldsymbol{\bar{v}}_i\right|\left|\boldsymbol{\bar{s}}_{i} \right|}\right)
    \end{split}
  \end{equation}

  \subsubsection{Total Loss}
  In summary, we define the total loss by combining the Eq. \ref{eq7} and Eq. \ref{eq8}:
  \begin{equation} \label{eq9}
    \begin{split}
      L=L_a+\beta L_s
    \end{split}
  \end{equation}
  where $\beta$ is a hyperparameter designed for the trade-off of the two loss terms. 
  While calculating the loss function ${L}$, it is important to note that some operations involved in the process are non-differentiable, such as sampling the probability distribution to retrieve words. 
  Thus, to ensure the update of the gradient, following~\citep{RennieMMRG17,00010KDBB22}, we optimize the model with REINFORCE algorithm~\citep{Williams92} with a self-critical baseline. 
  \begin{equation} \label{eq10}
    \begin{split}
      L=-\mathbb{E}\left[r\left(\Delta \mathrm{V}_i, \Delta \mathrm{S}_i\right) + \beta r\left(\boldsymbol{\bar{v}}_i, \boldsymbol{\bar{s}}_{i}\right)\right]
    \end{split}
  \end{equation}
  where $r\left(a_i, b_i\right) = \mathbb{E} \sum_{i=1}^n\left(\frac{a_i \cdot b_i}{\left|a_i \right|\left|b_i\right|}\right)$ is the score of attribute and semantic consistency. 
  The gradient of ${L}$ can be approximated as follows: 
  \begin{equation} \label{eq11}
    \begin{split}
      \nabla_\theta &L  \approx \\
      & \left(r\left(\Delta \mathrm{V}_i, \Delta \mathrm{S}_i^{{s}}\right)-r\left(\Delta \mathrm{V}_i, \Delta \mathrm{S}_i^g\right)\right) 
      \nabla_\theta \log P_\theta\left(\boldsymbol{s}_i^s \mid \boldsymbol{x}_i\right) \\
      & + \left(r\left(\boldsymbol{\bar{v}}_i, \boldsymbol{\bar{s}}_{i}^{{s}}\right)-r\left(\boldsymbol{\bar{v}}_i, \boldsymbol{\bar{s}}_{i}^g\right)\right)
      \nabla_\theta \log P_\theta\left(\boldsymbol{s}_i^s \mid \boldsymbol{x}_i\right) 
    \end{split}
  \end{equation}
  where $\boldsymbol{s}_i^s$ is a sampled caption, $r\left(a_i, b_i^{{s}}\right)$ and $r\left(a_i, b_i^g\right)$ define the samplely decoded score and the greedily decoded score obtained from the current model, respectively. 
  
  

  \section{Experiments} \label{sec:experiments}
  \begin{table*}[t]
    \setlength{\tabcolsep}{0.5mm}{
    \caption{Comparison with Multimodal Large Language Models (MLLMs) on CUB-200 and Oxford-102 datasets.}
    \label{tab:vlms}
    \begin{tabular}{l cccccccccccc}
    \toprule
    & \multicolumn{7}{c}{\emph{Supervised Metrics}} & \multicolumn{4}{c}{\emph{Diversity Metrics}} & \multicolumn{1}{c}{\emph{Unsupervised Metric}} \\
    \cmidrule(lr){2-8}
    \cmidrule(lr){9-12}
    \cmidrule(lr){13-12}
    Method & B@1 & B@2 & B@3 & B@4 & METEOR & ROUGE-L & CIDEr & Vocab &$\%$Novel & Length &\%Unique & CLIP-S \\
    \midrule
    \textbf{CUB-200} & \\
    ZeroCap & 9.5 & 1.8 & 0.4 & 0.1 & 3.9 & 11.5 & 1.9 & 50,257 & 100$\%$ & 8.8 & 96.6$\%$ & 76.4 \\
    Flamingo\_9B & 20.9 & 8.6 & 2.9 & 1.6 & 9.0 & 21.6 & 7.3 & 32,000 & 100$\%$ & 10.8 & 73.7$\%$ & 79.4 \\
    BLIP2\_2.7B & 18.5 & 6.8 & 2.8 & 1.2 & 8.5 & 20.8 & 6.8 & 50,272 & 100$\%$ & 11.2 & 74.4$\%$ & 78.9 \\
    BLIP2\_6.7B & 22.9 & 9.0 & 3.9 & 1.7 & 9.8 & 22.4 & 8.1 & 50,272 & 100$\%$ & 11.2 & 73.9$\%$ & \textbf{81.1} \\
    MiniGPT4\_7B & 9.2 & 3.0 & 1.9 & 0.0 & 9.8 & 17.0 & 0.0 & 32,000 & 100$\%$ & 92.3 & \textbf{99.8$\%$} & 56.6 \\
    InstructBLIP\_7B & 8.4 & 2.9 & 1.2 & 0.0 & 9.3 & 16.5 & 0.0 & 32,000 & 100$\%$ & \textbf{95.3} & 99.6$\%$ & 56.8 \\
    GeneIC & \textbf{24.3} & \textbf{11.4} & \textbf{5.8} & \textbf{3.1} & \textbf{11.0} & \textbf{24.3} & \textbf{20.1} & \textbf{50,272} & \textbf{100$\%$} & 18.6 & 81.2$\%$ & 79.7 \\
    \midrule
    \textbf{Oxford-102} & \\
    ZeroCap & 13.5 & 2.8 & 0.4 & 0.0 & 5.3 & 11.0 & 3.2 & 50,257 & 100$\%$ & 9.3 & 96.1$\%$ & 59.8 \\
    Flamingo\_9B & 23.2 & 7.4 & 2.7 & 1.0 & 10.9 & 18.2 & 13.3 & 32,000 & 100$\%$ & 10.9 & 52.3$\%$ & 78.5 \\
    BLIP2\_2.7B & 22.5 & 7.0 & 2.6 & 1.0 & 10.2 & 17.9 & 12.9 & 50,272 & 100$\%$ & 11.2 & 68.2$\%$ & 78.4 \\
    BLIP2\_6.7B & 23.4 & 7.1 & 2.7 & 1.1 & 10.8 & 18.7 & 14.8 & 50,272 & 100$\%$ & 11.4 & 51.2$\%$ & 79.1 \\
    MiniGPT4\_7B & 7.3 & 2.8 & 1.1 & 0.0 & \textbf{11.6} & 12.4 & 0.0 & 32,000 & 100$\%$ & 94.7 & \textbf{97.6$\%$} & 53.4 \\
    InstructBLIP\_7B & 8.5 & 2.5 & 0.8 & 0.0 & 11.2 & 12.3 & 0.0 & 32,000 & 100$\%$ & \textbf{97.5} & 97.5$\%$ & 55.4 \\
    GeneIC & \textbf{24.2} & \textbf{7.6} & \textbf{3.0} & \textbf{1.3} & 11.1 & \textbf{19.0} & \textbf{15.6} & \textbf{50,272} & \textbf{100$\%$} & 13.7 & 76.3$\%$ & \textbf{79.6} \\
    \bottomrule
    \end{tabular}}
  \end{table*}
  \subsection{Datasets} 
  In this paper, we compare two types of models aimed at improving the generalization ability on image captioning tasks: multimodal large language models (MLLMs)~\citep{TewelSSW22,AlayracDLMBHLMM22,li2023blip2,abs-2304-10592,abs-2305-06500} and cross-domain image captioning models~\citep{ChenLCHFS17,0007ZXFZ0L19,ZhaoXYYZFQ17,perdana2019instance,ZhaoWL21}. 
  MLLM is trained on web-scale datasets, including MSCOCO~\citep{LinMBHPRDZ14}, Visual Genome~\citep{KrishnaZGJHKCKL17}, CC3M~\citep{SoricutDSG18}, CC12M~\citep{Chen0CPPSGGMB0P23}, SBU~\citep{OrdonezKB11}, and LAION400M dataset~\citep{abs-2111-02114}, and has achieved remarkable zero-shot ability on different downstream tasks. 
  Cross-domain method utilizes the MSCOCO dataset~\citep{LinMBHPRDZ14} as the source domain and combines a limited quantity of data from a single target domain for joint training, thereby amplifying the model's performance within the target domain. 
  This includes Dual learning~\citep{ZhaoXYYZFQ17}, Multi-task~\citep{0007ZXFZ0L19}, Instance~\citep{perdana2019instance}, Retrieval~\citep{ZhaoWL21}, SCIC~\citep{YuanZHZXLW22} and LSML~\citep{ren2023crossing}. Among them, LSML explore domain generalization for image captioning, and our method follows a similar setup.

  In order to conduct a quantitative comparison between GeneIC and the comparison methods, we selected CUB-200~\citep{ReedALS16} and Oxford-102~\citep{NilsbackZ08} as the target domains due to their substantial domain shifts when compared to the source domain data. 
  More specifically: 
  \paragraph{CUB-200} It consists of 11,788 bird images from 200 different categories, each with 10 caption annotations. We followed the data splitting method described in~\citep{YuanZHZXLW22} and selected 5,788 images as the test set. 
  
  \paragraph{Oxford-102} It comprises 8,189 flower images distributed across 102 categories, and each image contains 10 captions. Our preferred data splitting method, outlined in~\citep{YuanZHZXLW22}, guided our selection of 1,000 images for the test set. 
  
  In addition, we also used more images with large domain shift to qualitatively demonstrate the effectiveness of our method, including Food101~\citep{BossardGG14}, StanfordCars~\citep{Krause0DF13}, which are not annotations. 
  We randomly selected 1,000 images as the target domain test set. 
  These dataset cover different scenarios including animals, plants, machines, and food, forming a comprehensive evaluation benchmark.

  \subsection{Training Details}
  
  GeneIC is a universal framework. 
  To demonstrate the superiority of this framework, we selected the current state-of-the-art zero-shot image captioning method as the backbone, namely BLIP2~\citep{li2023blip2}, and utilizing OPT2.7B as the decoder. 
  Meanwhile, the number of learnable prompt vectors in GeneIC is set to $M=8$, and the training image count amounts to $N=1,000$. 
  The learnable prompt vectors of GeneIC are randomly initialized by drawing from a Gaussian distribution with zero-mean and standard deviation equal of 0.02. 
  In Formula~\ref{eq9}, the hyperparameter $\beta$  is set to 0.5. 
  The model is built upon the open-source code of BLIP2. \footnote{\url{https://github.com/salesforce/LAVIS/tree/main/projects/blip2}.}
  During the training process, we use AdamW~\citep{LoshchilovH19} as the optimizer and set the epoch and batchsize to 30 and 10, with an initialized learning rate of $5 \times 10^{-4}$, which is decayed by the cosine annealing rule.

  \subsection{Baseline Methods and Evaluation Metrics}
  
  In this paper, we compare GeneIC with two baseline methods: Multimodal Large Language Models (MLLMs) and cross-domain image captioning models.  
  MLLMs achieve zero-shot generation by utilizing hand-crafted prompt. 
  Among them, ZeroCap~\citep{TewelSSW22} directly bridges the visual encoder and the language decoder without requiring additional training. 
  Flamingo\_9B~\citep{AlayracDLMBHLMM22}, BLIP2\_6.7B and BLIP2\_2.7B~\citep{li2023blip2} employ joint training of the visual encoder and the language decoder. 
  MiniGPT4\_7B~\citep{abs-2304-10592} and InstructBLIP\_7B~\citep{abs-2305-06500} perform additional fine-tuning of the models based on BLIP2 by utilizing meticulously crafted instruction datasets.  
  Adhering to the original settings of the baseline methods, for ZeroCap, Flamingo, BLIP2\_6.7B, and BLIP2\_2.7B, we employ ``A photo of'' as the hand-crafted prompt. 
  For MiniGPT4\_7B and InstructBLIP\_7B, we used ``Describe this image in detail." as the hand-crafted prompt.
  Further relevant details will be discussed in Appendix~\ref{appx:more_details}. 
  Cross-domain methods entail joint training on the source domain data ($i.e.$, MSCOCO) along with a limited amount of target domain data, enabling cross-domain image captioning. 
  Due to the unavailability of source code for certain baseline methods, we solely present the result provided in the original papers of those methods. 
  
  This paper examines the quality of generated sentences through three dimensions: supervised metrics, diversity metrics, and unsupervised metrics. 
  Supervised metrics encompass conventional evaluation metrics for image captioning, such as BLEU~\citep{PapineniRWZ02}, METEOR~\citep{BanerjeeL05}, ROUGE-L~\citep{lin-2004-rouge}, and CIDEr~\citep{VedantamZP15}. 
  These metrics assess the degree of alignment between generated sentences and ground truth, utilizing n-gram methods. 
  Due to the inclusion of fine-grained target domain object descriptions in human annotations, such as the color of birds, it becomes essential to employ supervised metrics directly for evaluating the accuracy of the generated sentences. 
  Diversity metrics quantify the diversity exhibited by generated sentences. Following~\citep{TewelSSW22}, in this paper, Vocab, \%Novel, and Length are employed as indicators to assess the model's capacity for diversity. 
  Specifically, Vocab corresponds to the vocabulary size, \%Novel denotes none of the generated sentences appear in the training set, and Length represents the average sentence length. 
  Furthermore, we define \%Unique as the proportion of generated sentences without any repetition. 
  Unsupervised metrics directly capture the similarity between input images and generated sentences. 
  A typical metric, CLIP-S~\citep{00010KDBB22}, quantifies the cosine similarity between features of images and sentence extracted from pretrained CLIP model. 

  \begin{table*}[!ht]
    \caption{Comparison with Cross-domain methods on CUB-200 and Oxford-102 datasets. ``-'' represents the results have not been given in the raw paper.}
    \label{tab:cross-domains}
    \setlength{\tabcolsep}{0.7mm}
    \begin{tabular}{l cccccccccccc}
    \toprule
    & \multicolumn{7}{c}{\emph{Supervised Metrics}} & \multicolumn{4}{c}{\emph{Diversity Metrics}} & \multicolumn{1}{c}{\emph{Unsupervised Metric}} \\
    \cmidrule(lr){2-8}
    \cmidrule(lr){9-12}
    \cmidrule(lr){13-12}
    Method & B@1 & B@2 & B@3 & B@4 & METEOR & ROUGE-L & CIDEr & Vocab &\%Novel & Length &\%Unique & CLIP-S \\
    \midrule
    \textbf{CUB-200} & \\
    Dual learning & - & - & - & - & - & - & - & - & - & - & - & - \\
    Multi-task & 92.3 & 83.2 & 70.6 & 57.5 & 37.4 & \textbf{72.0} & 77.3 & 812 & 73.5$\%$ & 9.5 & 58.5$\%$ & 78.2 \\
    Instance & 90.9 & 81.2 & 53.2 & 32.9 & 27.9 & 58.9 & 25.9 & - & - & - & - & - \\
    Retrieval & \textbf{95.3} & \textbf{83.9} & \textbf{72.0} & \textbf{61.6} & 36.6 & 69.3 & 76.7 & - & - & - & - & - \\
    SCIC & 93.8 & 81.6 & 71.4 & 61.1 & \textbf{36.9} & 70.7 & \textbf{78.2} & - & - & - & - & - \\
    \midrule
    LSML & 20.4 & 8.0 & 3.2 & 1.3 & 10.2 & 20.9 & 9.6 & - & - & - & - & - \\
    GeneIC & 24.3 & 11.4 & 5.8 & 3.1 & 11.0 & 24.3 & 20.1 & \textbf{50,272} & \textbf{100$\%$} & \textbf{18.6} & \textbf{81.2$\%$} & \textbf{79.7} \\
    \midrule
    \textbf{Oxford-102} & \\
    Dual learning & 91.2 & 84.4 & 77.1 & 71.6 & 43.0 & 82.4 & 79.7 & - & - & - & - & - \\
    Multi-task & 91.0 & 83.8 & 78.4 & 72.1 & {45.3} & {82.9} & 89.2 & 1,509 & 70.2$\%$ & 10.2 & 67.7$\%$ & 77.4 \\
    Instance & 85.9 & 77.2 & 67.9 & 61.1 & 36.5 & 72.9 & 29.2 & - & - & - & - & - \\
    Retrieval & \textbf{96.6} & \textbf{91.8} & \textbf{86.0} & \textbf{80.2} & 42.2 & 77.8 & 87.3 & - & - & - & - & - \\
    SCIC & 92.7 & {85.4} & {78.9} & {74.1} & \textbf{46.6} & \textbf{84.9} & \textbf{90.8} & - & - & - & - & - \\
    \midrule
    LSML & 19.4 & 5.9 & 2.2 & 0.9 & 9.7 & 18.0 & 14.4 & - & - & - & - & - \\
    GeneIC & 24.2 & 7.6 & 3.0 & 1.3 & 11.1 & 19.0 & 15.6 & \textbf{50,272} & \textbf{100$\%$} & \textbf{13.7} & \textbf{76.3$\%$} & \textbf{79.6} \\
    \bottomrule
    \end{tabular}
  \end{table*}

  \subsection{Quantitative Analysis}
  \begin{table*}[t]
    \centering
    \caption{Comparison with Multimodal Large Language Models (MLLMs) on Food101 and StanfordCars datasets.}
    \scalebox{1.0}{
      \label{tab:car_food}  
    \begin{tabular}{llccccccc}  
      \toprule 
      &  & \multicolumn{7}{c}{\emph{StanfordCars dataset}} \\
      \cmidrule(lr){3-9}
      &  & ZeroCap & Flamingo\_9B & BLIP2\_2.7B & BLIP2\_6.7B & MiniGPT4\_7B & InstructBLIP\_7B & GeneIC \\
      \midrule      
      & CLIP-S & 73.9 & 78.5 & 77.9 & \textbf{79.4} & 50.8 & 51.0 & 79.1 \\
      \midrule
      &  & \multicolumn{7}{c}{\emph{Food101 dataset}} \\
      \cmidrule(lr){3-9}
      &  & ZeroCap & Flamingo\_9B & BLIP2\_2.7B & BLIP2\_6.7B & MiniGPT4\_7B & InstructBLIP\_7B & GeneIC \\
      \midrule  
      & CLIP-S & 71.9 & 80.3 & 79.8 & 80.6 & 52.7 & 53.3 & \textbf{81.1} \\
      \bottomrule
    \end{tabular}}
  \end{table*}
  \begin{figure*}[!t]
    \centering
    \includegraphics[width=.98\textwidth]{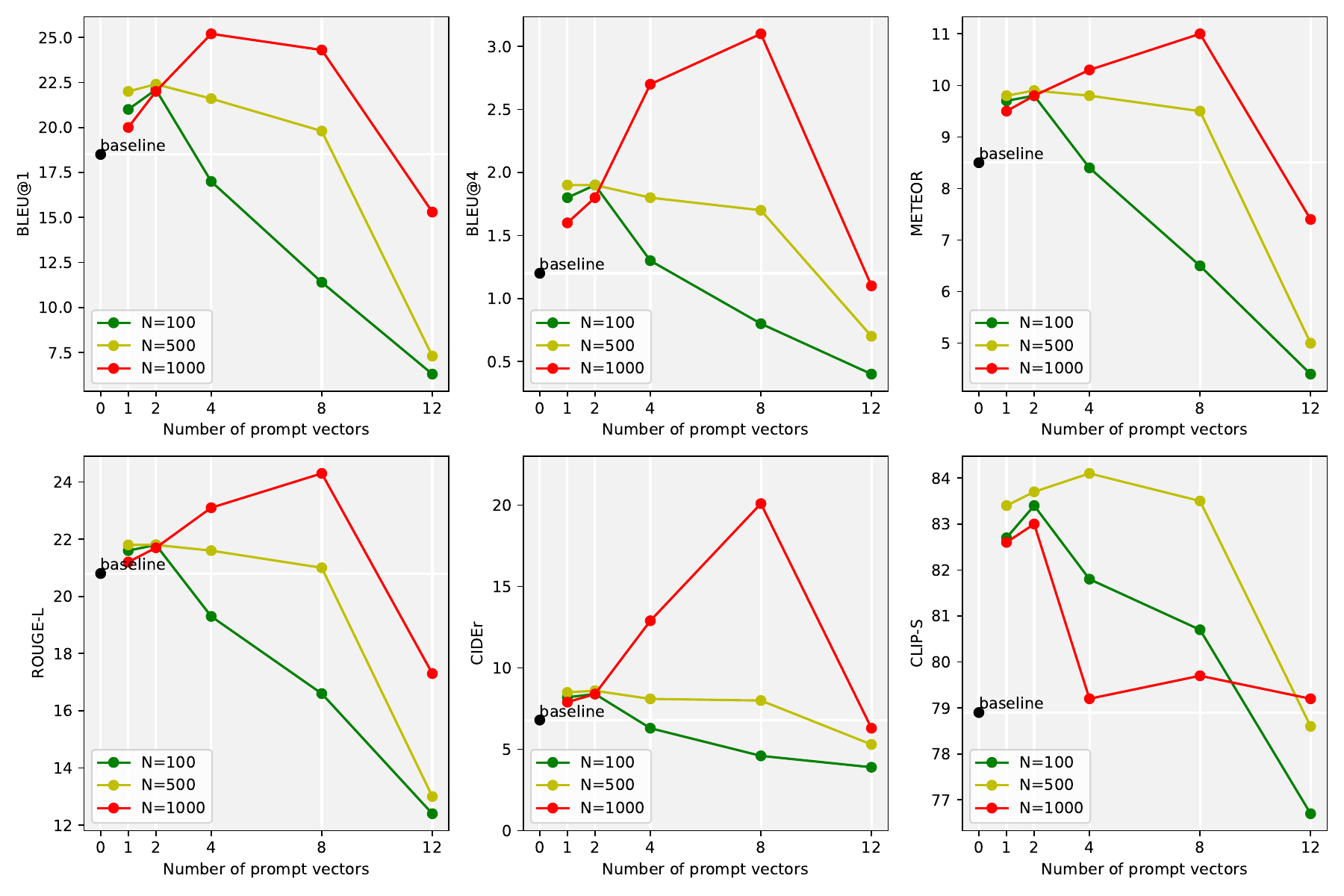}
    \caption{Investigations on the number of prompt vectors and training images. $N$ denotes number of training images. The black dots represent the results of the baseline method BLIP2\_2.7B.  }
    \label{fig:main_results}
  \end{figure*}
  \paragraph{Comparison with MLLMs}
  
  Table~\ref{tab:vlms} reports the results of GeneIC and MLLMs on the CUB-200 and Oxford-102 datasets. 
  GeneIC's default settings are $M=8$ and $N=1000$, where $M$ indicates the number of learnable prompt vectors, and $N$ represents the number of target domain images. 
  The results show that 1) ZeroCap underperforms other methods on most metrics, indicating that merely bridge single modal visual and language models is insufficient for effectively aligning visual and language modalities, and it lags behind multimodal methods trained on image-text pairs. 
  2) GeneIC exhibits superiority over Flamingo\_9B and BLIP2\_6.7B in supervised metrics, despite the latter having larger model sizes. 
  Specifically, when compared to BLIP2\_6.7B, GeneIC achieves improvements of 12.0 and 0.8 on the CIDEr metrics, respectively, for the CUB-200 and Oxford-102 dataset. 
  This phenomenon suggests that hand-crafted prompt ($i.e.$,``A photo of'') cannot guide the model adaptively to focus on domain-specific attributes in the target domain. 
  Hand-crafted prompt direction fail to enable generalization of the model to target domains with significant domain shifts from the source domain. 
  3) Regarding diversity metrics, GeneIC generates longer sentences and higher \%Unique compared to Flamingo\_9B, BLIP2\_6.7B and BLIP2\_2.7B. 
  This phenomenon indicates that the sentences generated by GeneIC contain more informative content. 
  4) As for unsupervised metrics, on the CUB-200 dataset, GeneIC slightly outperforms Flamingo\_9B and BLIP2\_2.7B, but falls behind BLIP2\_6.7B. 
  This discrepancy that compared to supervised metrics can be attributed to the human annotations in the CUB-200 dataset, which primarily focus on describing birds while neglecting the surrounding environment. 
  Consequently, supervised metrics prioritize measuring the accuracy of the main objects in the target domain images. 
  In contrast, CLIP-S provides a comprehensive evaluation of all the content in the images. 
  GeneIC ensures the comprehensive generation of sentences while giving emphasis to the description of the main objects in the target domain images. 
  Conversely, on the Oxford-102 dataset, GeneIC achieves the best performance across all metrics. 
  This advantage stems from the fact that the images in the Oxford-102 dataset are usually close-ups of flowers. 
  5) MiniGPT4\_7B and InstructBLIP\_7B construct additional instruction datasets for fine-tuning the models, leading to the generation of sentences that different style from the ground-truth human annotations and are longer in length. 
  Consequently, they demonstrate inferior performance on most evaluation metrics. 
  To ensure a fair comparison, we provide the generated examples of MiniGPT4\_7B and InstructBLIP\_7B in Appendix~\ref{appx:more_vis}, accompanied by further discussions.
  However, it is noteworthy that, unlike MiniGPT4\_7B and InstructBLIP\_7B, GeneIC does not require any annotated data. 
  Additionally, during the training process, GeneIC freezes most of the model parameters and only optimizes the prompt vectors, significantly reducing training costs.
  \begin{table*}[t]
    \centering
    \caption{The performance of different loss terms on CUB-200 and Oxford-102 datasets.}
    \label{tab:ablation}
    \begin{tabular}{lcc|cccccc}  
      \toprule 
      & $L_a$ & $L_s$ & B@1 & B@4 & METEOR & ROUGE-L & CIDEr & CLIP-S \\
      \midrule      
      \multirow{3}*{\rotatebox[origin=c]{90}{\textbf{CUB}}}
      & \checkmark &  & \textbf{25.2} & 2.7 & 10.3 & 23.1 & 17.9 & 73.5 \\
      &  & \checkmark  & 21.6 & 1.8 & 9.8 & 21.6 & 8.1 & \textbf{84.1} \\
      & \checkmark & \checkmark  & 24.3 & \textbf{3.1} & \textbf{11.0} & \textbf{24.3} & \textbf{20.1}  & 79.7 \\
      \midrule      
      \multirow{3}*{\rotatebox[origin=c]{90}{\textbf{Oxford}}}
      & \checkmark &  & 23.7 & 1.0 & 10.3 & 18.3 & 14.7 & 79.0 \\
      &  & \checkmark  & 22.5 & 0.9 & 10.2 & 17.9 & 13.8 & 79.3 \\
      & \checkmark & \checkmark  & \textbf{24.2} & \textbf{1.3} & \textbf{11.1} & \textbf{19.0} & \textbf{15.6} & \textbf{79.6}  \\
      \bottomrule
    \end{tabular}
  \end{table*}

  \begin{table*}[!t]
    \centering
    \caption{The performance of different attribute-transferred image construction methods on CUB-200 and Oxford-102 datasets.}
    \label{tab:transfer_ablation}
    \begin{tabular}{l ccccccc}
    \toprule
    & & B@1 & B@4 & METEOR & ROUGE-L & CIDEr & CLIP-S \\
    \midrule
    \multirow{4}*{\rotatebox[origin=c]{90}{\textbf{CUB}}}
    & Random & 21.6 & 2.4 & 9.9 & 20.3 & 8.6 & 80.0 \\
    & Augmentation & 20.0 & 1.8 & 9.2 & 19.1 & 7.5 & 76.1 \\
    & Cluster & 22.3 & 2.6 & 10.7 & 21.3 & 11.3 & \textbf{80.6} \\
    & GeneIC & \textbf{24.3} & \textbf{3.1} & \textbf{11.0} & \textbf{24.3} & \textbf{20.1}  & 79.7 \\
    \midrule
    \multirow{4}*{\rotatebox[origin=c]{90}{\textbf{Oxford}}}
    & Random & 21.4 & 0.7 & 9.3 & 17.5 & 12.8 & 76.7 \\
    & Augmentation & 20.2 & 0.7 & 9.2 & 17.1 & 12.3 & 76.0 \\
    & Cluster & 22.5 & 1.1 & 9.9 & 18.4 & 14.2 & 78.9 \\
    & GeneIC & \textbf{24.2} & \textbf{1.3} & \textbf{11.1} & \textbf{19.0} & \textbf{15.6} & \textbf{79.6} \\
    \bottomrule
    \end{tabular}
  \end{table*}
  \paragraph{Comparison with Cross-domain Methods}
  
  Table~\ref{tab:cross-domains} presents the results of GeneIC and cross-domain methods on the CUB-200 and Oxford-102 datasets. 
  The results indicate that GeneIC performs worse than the comparison methods that utilize target domain data in terms of supervised metrics. 
  This result can be attributed to distinct language styles. 
  Evaluating the matching degree between generated sentences and ground truth within a specific word range is necessary for supervised metrics. 
  Therefore, in the presence of substantial differences in language styles, even if the generated sentences accurately convey the intended meaning, they may receive lower scores. 
  The comparison methods that utilize target domain data have a clear advantage in aligning language styles as they incorporate training data that closely resembles the ground truth annotations. 
  It is noteworthy that GeneIC exhibits significant advantages in terms of diversity metrics and unsupervised metrics. 
  This phenomenon indicates that the sentences generated by GeneIC exhibit greater diversity and comprehensiveness. 
  In contrast, when there is no target domain data available, GeneIC outperforms LSML in terms of supervised metrics. 
  This resulting demonstrates that GeneIC exhibits superior generalization capability. 

  \begin{table*}[t]
    \centering{
    \caption{The retrieval and generated words for each of the 8 prompt vectors learned by GeneIC, and the retrieval distance shown in parentheses. 
    N/A means non-Latin characters.}
    \label{tab:vis_ctx_vec}
    \setlength{\tabcolsep}{0.9mm}
    \begin{tabular}{c |rr|rr|rr|rr}
    \toprule
    & \multicolumn{2}{c}{{CUB-200}} & \multicolumn{2}{c}{{Oxford-102}} & \multicolumn{2}{c}{{StanfordCars}} & \multicolumn{2}{c}{{Food101}} \\
    \cmidrule(lr){2-3}
    \cmidrule(lr){4-5}
    \cmidrule(lr){6-7}
    \cmidrule(lr){8-9}
  
    \# & Retrieval & Generate & Retrieval & Generate & Retrieval & Generate & Retrieval & Generate \\
    \midrule
    1 & N/A (1.3865) & a & Honestly (1.4644) & a & Honestly (1.4595) & a & N/A (1.4939) & a \\
    2 & N/A (1.4178) & a & Honestly (1.4610) & a & N/A (1.4609) & a & N/A (1.4715) & a \\
    3 & N/A (1.4195) & bird & N/A (1.4729) & a  & N/A (1.4474) & a & Honestly (1.4701) & a \\
    4 & Yeah (1.4302) & a & N/A (1.4172) & flower & Honestly (1.4198) & car & N/A (1.4694) & dish \\
    5 & Honestly (1.4128) & a & Yeah (1.4910) & a  & Despite (1.4326) & a & N/A (1.4576) & of \\
    6 & N/A (1.3905) & a & Despite (1.4985) & the  & N/A (1.4140) & a & Despite (1.4478) & sc \\
    7 & N/A (1.3928) & A & N/A (1.5002) & a & N/A (1.4113) & a & Yeah (1.4336) & of \\
    8 & N/A (1.3747) & the & N/A (1.4546) & is  & Yeah (1.4112) & a & N/A (1.5096) & of \\
    \bottomrule
    \end{tabular}}
  \end{table*}

  \begin{figure*}[!t]
    \centering
    \includegraphics[width=.98\textwidth]{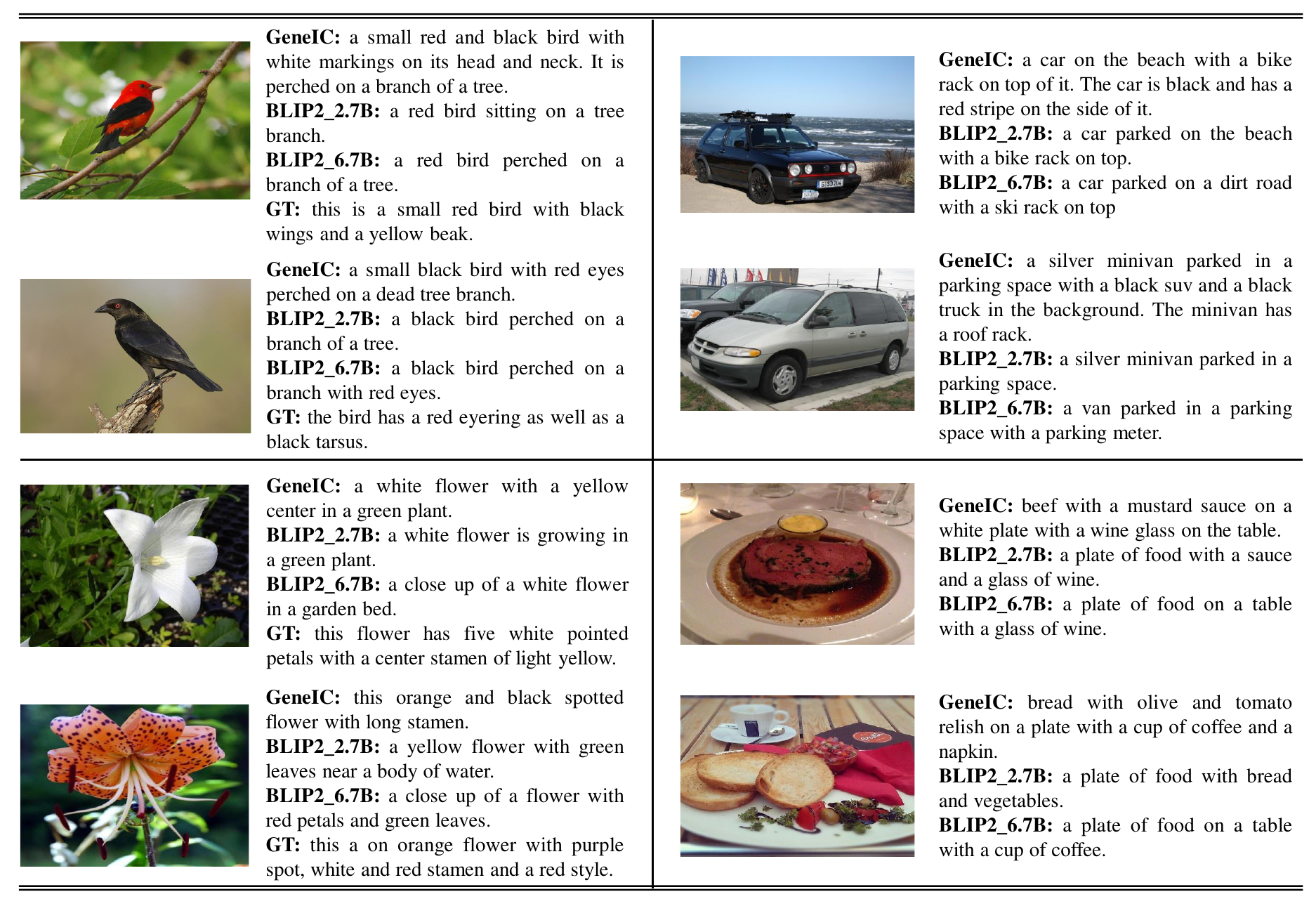}
    \caption{Examples of captions generated by GeneIC and baseline models as well as the corresponding ground truth (GT is one of the 10 given annotated captions).}
    \label{fig:case}
  \end{figure*}

  \paragraph{The Results on Food101 and StanfordCars Datasets}
  
  Table~\ref{tab:car_food} only reports the unsupervised metrics results of GeneIC and MLLMs on the StanfordCars and Food101 datasets, as these datasets lack ground-truth annotations. 
  Cross-domain methods cannot directly generalize to target domain data without human annotations, as they are required during the training process. 
  This limitation hampers the applicability of cross-domain methods. 
  In contrast, GeneIC demonstrates the ability to generalize to any target domain with only a small number of images. 
  Moreover, the results in Table~\ref{tab:car_food} demonstrate that GeneIC achieves similar result to that in Table~\ref{tab:vlms} on the StanfordCars and Food101 datasets. 
  On the StanfordCars dataset, GeneIC slightly underperforms BLIP2\_6.7B. 
  However, on the Food101 dataset, GeneIC achieves the best results. 
  This is because the StanfordCars dataset contains more background information, while the Food101 dataset typically focuses on close-up shots of food. 

  \subsection{Influence of Prompt Vectors Length and Number of Training Images}
  To examine the influence of prompt vectors length and the number of target domain images on the model, we conducted experiments with various parameters using the CUB-200 dataset. 
  Specifically, we concurrently varied the number of prompt vectors $M=\{1, 2, 4, 8, 12\}$ and the number of training images $N=\{100, 500, 1000\}$. 
  The results are depicted in Figure~\ref{fig:main_results}, where the black dots represent the results of the baseline method BLIP2\_2.7B.

  In terms of supervised metrics, GeneIC demonstrates a trend of performance that initially increases and then decreases. 
  Taking the CIDEr metric as an example, when the $M=1$, GeneIC trained on different quantities of target domain image sets yield similar scores. 
  This phenomenon arises from the inability of shorter prompt vectors to capture complex semantics. 
  Consequently, when the $M=2$, the model demonstrates a similar trend. 
  Substantial variations in model performance are observed across diverse training data configurations as the number of prompt vectors continues to increase, specifically at $M=4$. 
  As the quantity of training data increases, performance improves accordingly. 
  Nevertheless, when confronted with longer vector lengths and limited training data, the model's performance may deteriorate further, falling below the level achieved by models utilizing shorter vectors. 
  This phenomenon implies that longer prompt vectors necessitate a larger amount of training data to converge. 
  It is noteworthy that with $M=1$, an increase in training data actually results in decreased performance. 
  We believe this phenomenon to the limited capability of shorter vectors in capturing the diverse attributes of the target domain dataset. 
  When $M=12$, the model exhibits the poorest performance across all configurations. 
  This can be attributed to two factors. 
  Firstly, the excessively long prompt vectors introduce variable-length sequences that pose challenges to the autoregressive model, deviating from its pre-training process. 
  Secondly, the prompt vectors suffers from overfitting to the data.
  
  On the unsupervised metric CLIP-S, we observed a decrease in model performance when $N=1,000$, in contrast to the results obtained from the supervised metrics. 
  This discrepancy can be attributed to the nature of CLIP-S, which evaluates the similarity between generated sentences and all elements present in the input images. 
  In contrast, the supervised metrics primarily evaluate the correlation between generated sentences and main objects in the images, such as birds and flowers, disregarding the image background. 
  This is determined by the ground-truth annotation. 
  Consequently, as the model undergoes extensive training, the performance of CLIP-S inevitably diminishes, necessitating a trade-off between comprehensiveness and specificity. 
  Nonetheless, it is noteworthy that GeneIC consistently surpasses the performance of the baseline methods. 

  \subsection{Ablation Study} 
  To validate the effectiveness of each introduced module in this paper, we conducted comprehensive ablation experiments on the CUB-200 and Oxford-102 datasets. 
  
  \paragraph{Semantic and Attribute Consistency}
  Table~\ref{tab:ablation} presents the ablation results of different loss terms. 
  The findings reveal that $L_a$ demonstrates superior performance on supervised metrics, while $L_s$ outperforms in unsupervised metrics. 
  This disparity can be attributed to the bias present in the ground-truth annotations, which primarily emphasizes describing target domain objects in the images ($e.g.$, birds and flowers), while disregarding the background. 
  Conversely, the unsupervised metric CLIP-S equally considers all contents within the images. 
  By integrating both loss terms, GeneIC ensures that the generated sentences maintain a focus on target domain objects while also considering other elements in the images.

  \paragraph{Attribute Transfer of Intra-Domain Images}
  Table~\ref{tab:transfer_ablation} illustrates the impact of different attribute-transferred image construction methods on the model's performance. 
  In this context, ``Random'' signifies the random selection of an image as the attribute-transferred image to calculate $L_a$. 
  ``Augmentation'' involves the generation of attribute-transferred images through data augmentation, with this paper employing the ``cutmix'' augmentation method~\citep{YunHCOYC19}. 
  ``Cluster'' indicates the utilization of CLIP retrieval to select the most similar image from the training set as the attribute-transferred image. 
  The findings reveal that method ``Augmentation'' yields the poorest results across all evaluation metrics. 
  This can be attributed to the fact that the augmented images do not conform to reality, depicting situations such as birds without heads or forests appearing in the sea, consequently leading to negative implications on the optimization of $L_s$. 
  However, GeneIC demonstrates a notable enhancement in supervised metrics, implying that images generated through attribute transfer can effectively guide the model's focus towards domain-specific knowledge. 

  \subsection{Interpretability of Learned Prompt Vectors}
  We adopt two methods, retrieval, and generation, to interpret the learned prompts. 
  Following~\citep{ZhouYLL22}, as the prompt vectors are learned in a continuous space, an possible approach is to retrieve the word in the vocabulary closest to the prompt vector based on Euclidean distance. 
  Simultaneously, considering the backbone as a generative model, we directly input the prompt vector into the model to generate the corresponding word (selecting the word with the highest probability as the final output). 
  The retrieval and generation result on four datasets are presented in Table~\ref{tab:vis_ctx_vec}. 
  It is observed that, across different datasets, the retrieval results are often similar, frequently containing words such as ``Yeah'', ``Honestly'', and ``Despite'', where ``Despite'' is somewhat relevant to image captioning. 
  However, when all the words are concatenated, the prompts lose coherence. 
  This finding aligns with the conclusion from~\citep{ZhouYLL22}. 
  It is important to note that drawing definite conclusions solely based on the retrieved results may be inaccurate, as explaining the learned prompts using the nearest words can be misleading—the semantic meaning of vectors may not necessarily correlate with the closest words. 
  As for the generation method, we discover that the prompt vectors manifest distinct domain-specific knowledge. 
  For instance, words such as ``bird'', ``flower'', ``car'', and ``dish'' accurately reflect the main objects in the target domain. 
  This underscores the prompt vectors' effective learning of domain-specific knowledge. 

  \begin{table*}[t]
    \centering
    \caption{Comparison with more prompts on CUB-200 and Oxford-102 datasets using the same backbone.}
    \label{tab:prompt}
    \begin{tabular}{l ccccccc}
    \toprule
    &Prompt & B@1 & B@4 & METEOR & ROUGE-L & CIDEr & CLIP-S \\
    \midrule
    \multirow{5}*{\rotatebox[origin=c]{90}{\textbf{CUB-200}}}
    & ``A photo of'' & 18.5 & 1.2 & 8.5 & 20.8 & 6.8 & 78.9  \\
    & ``A bird photo depict'' & 11.9 & 0.5 & 6.7 & 18.6 & 3.4 & \textbf{80.6} \\
    & ``Describe this image in detail'' & 13.0 & 0.5 & 7.1 & 18.1 & 3.8 & 77.9 \\
    & Randomly prompt vectors & 7.9 & 0.4 & 5.5 & 15.1 & 2.8 & 79.4 \\
    & GeneIC & \textbf{24.3} & \textbf{3.1} & \textbf{11.0} & \textbf{24.3} & \textbf{20.1}  & 79.7 \\
    \midrule
    \multirow{5}*{\rotatebox[origin=c]{90}{\textbf{Oxford-102}}}
    & ``A photo of'' & 22.5 & 1.0 & 10.2 & 17.9 & 12.9 & 78.4 \\
    & ``A flower photo depict'' & 13.9 & 0.7 & 9.6 & 18.2 & 12.6 & 78.3 \\
    & ``Describe this image in detail'' & 14.2 & 0.7 & 9.1 & 17.6 & 12.8 & 78.4 \\
    & Randomly prompt vectors & 12.5 & 0.4 & 7.9 & 16.3 & 8.8 & 77.9 \\
    & GeneIC & \textbf{24.2} & \textbf{1.3} & \textbf{11.1} & \textbf{19.0} & \textbf{15.6} & \textbf{79.6}  \\
    \bottomrule
    \end{tabular}
  \end{table*}
  
  \begin{table*}[t]
    \centering
    \caption{Compare universal and specific prompt vectors using the same backbone.}
    \label{tab:change_vec}
    \begin{tabular}{l cccccc}
    \toprule
    Dataset & B@1 & B@4 & METEOR & ROUGE-L & CIDEr & CLIP-S \\
    \midrule
    \textbf{Hand-crafted prompt} & \\
    CUB-200 & 18.5 & 1.2 & 8.5 & 20.8 & 6.8 & 78.9 \\
    Oxford-102 & 22.5 & 1.0 & 10.2 & 17.9 & 12.9 & 78.4 \\
    Food101 & - & - & - & - & - & 77.9 \\
    StanfordCars & - & - & - & - & - & 79.8 \\
    \midrule
    \textbf{CUB-200 prompt vectors} & \\
    CUB-200 & 24.3 & 3.1 & 11.0 & 24.3 & 20.1  & 79.7 \\
    Oxford-102 & 14.7 & 0.4 & 5.4 & 11.3 & 4.2 & 79.2 \\
    Food101 & - & - & - & - & - & 79.0 \\
    StanfordCars & - & - & - & - & - & 78.6 \\
    \midrule
    \textbf{Domain-specific prompt vectors} & \\
    CUB-200 & 24.3 & 3.1 & 11.0 & 24.3 & 20.1  & 79.7 \\
    Oxford-102 & 24.2 & 1.3 & 11.1 & 19.0 & 15.6 & 79.6  \\
    Food101 & - & - & - & - & - & 79.1 \\
    StanfordCars & - & - & - & - & - & 81.1 \\
    \bottomrule
    \end{tabular}
  \end{table*}
  
  \begin{figure*}[t]
    \centering
    \includegraphics[width=.91\textwidth]{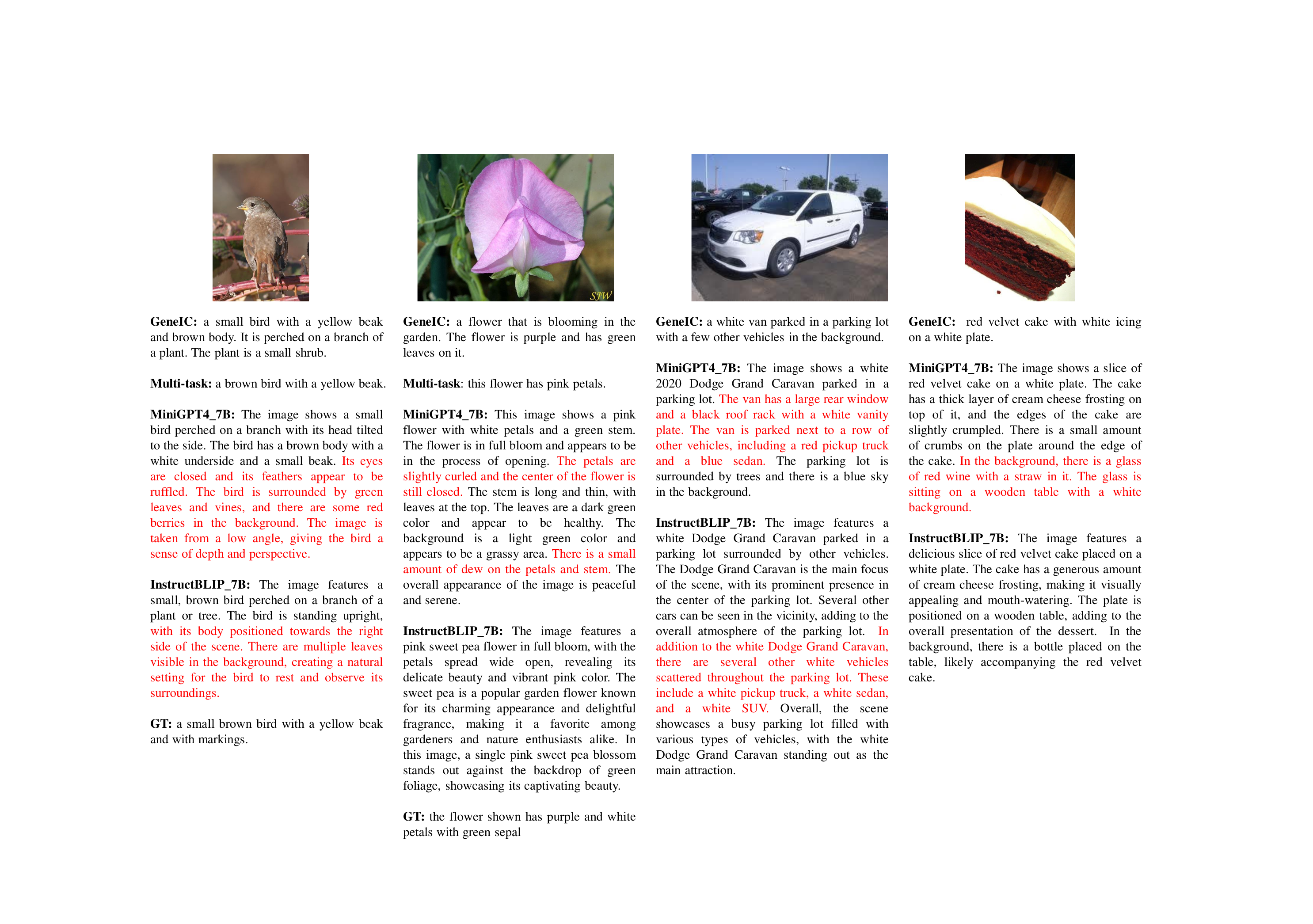}
    \caption{More examples of captions generated by GeneIC and comparison methods as well as the corresponding ground truth (GT is one of the 10 given annotated captions). 
    Red represents hallucination.}
    \label{fig:case_app}
  \end{figure*}

  \begin{figure*}[!t]
    \centering
    \includegraphics[width=.78\textwidth]{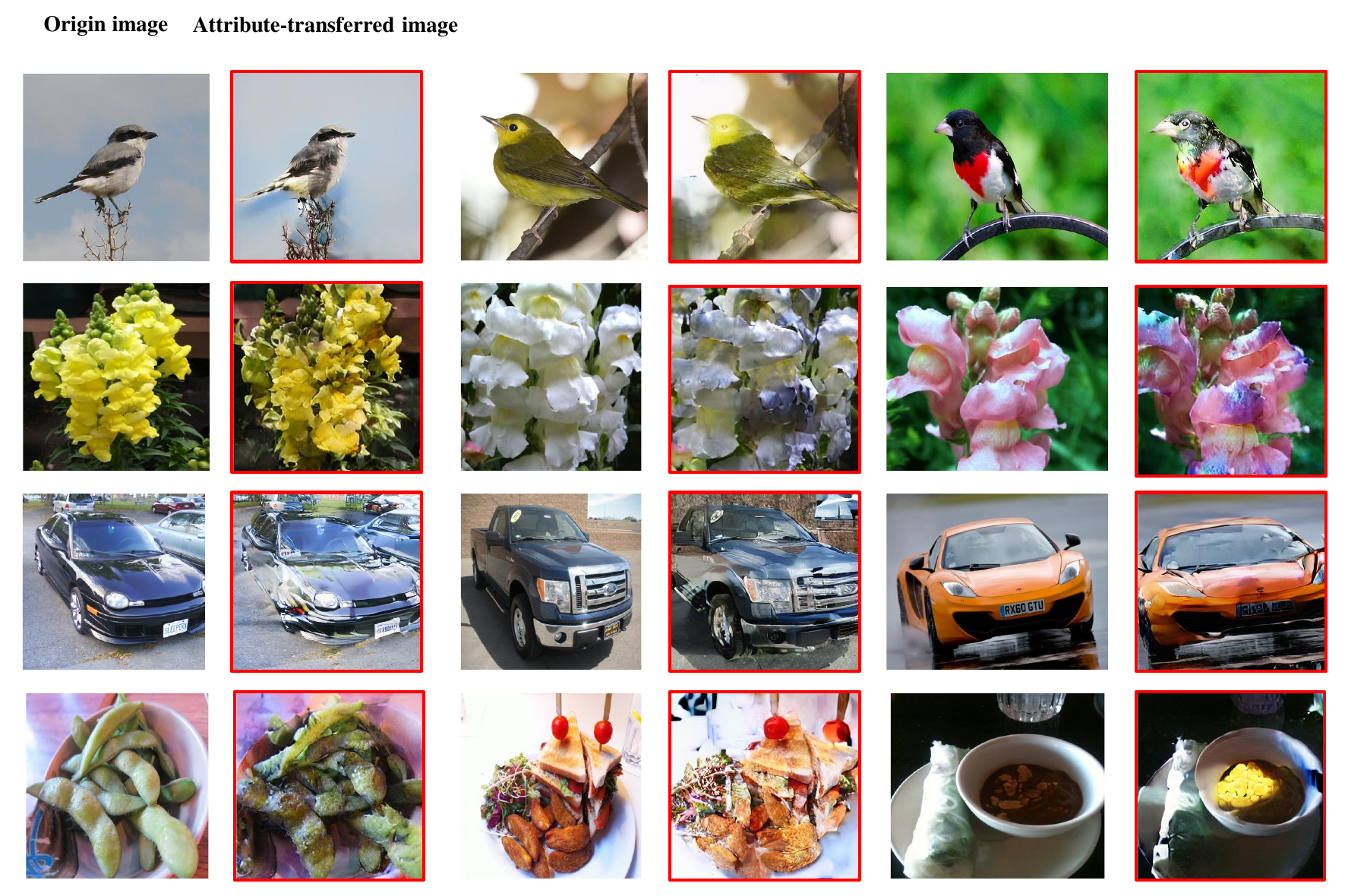}
    \caption{Examples of images generated by attribute transfer of intra-domain images.}
    \label{fig:case_attr}
  \end{figure*}

  \subsection{Visualization and Analysis} \label{sec:vis_analysis}
  
  Figure~\ref{fig:case} presents examples generated by GeneIC and the baseline method BLIP2 across four datasets. 
  It is noteworthy that cross-domain methods necessitate training the model with annotated data from the target domain and are not applicable to scenarios lacking annotations. 
  Furthermore, both MiniGPT4\_7B and InstructBLIP\_7B utilize additional supervised data for fine-tuning the models. 
  A comprehensive analysis of these two methods will be provided in the Appendix~\ref{appx:more_vis}.
  In these examples, we observe that the baseline method follows a consistent mode, characterized by ``a + object + background.'' 
  This mode arises due to the use of hand-crafted prompts, which lead to a collapse of diversity in generated sentences and a lack of domain-specific knowledge. 
  In contrast, our approach addresses this issue by adopting specific prompt vectors for each target domain, thereby guiding the model to focus on domain-specific knowledge and generate more informative descriptions.

  \section{Concluding Remarks} \label{sec:conclusion} 
  With the development of Large Language Models (LLMs), an increasing number of researchers are turning their attention to enabling LLM to process visual inputs, giving rise to a series of Multi-Modal Large Language Models (MLLMs). 
  By utilizing hand-crafted prompts, these models have achieved remarkable zero-shot performance across different downstream tasks, including image captioning. 
  However, when confronted with significant domain shifts, the utilization of hand-crafted prompts results in sentences with similar modes, a phenomenon known as mode collapse. 
  Consequently, this limitation hampers diversity and the incorporation of domain-specific knowledge in the generated sentences. 
  To address this issue, some studies have introduced extensive instruction datasets for the purpose of fine-tuning the models. 
  Nevertheless, it is essential to acknowledge that the construction of instruction datasets comes with considerable costs. 
  Meanwhile, it may result hallucination.

  This paper introduces an unsupervised prompt learning method aimed at enhancing the model's generalization capability in image captioning without annotations. 
  The results demonstrate that, in comparison to hand-crafted prompts, this method optimizes the prompt vectors by acquiring domain-specific knowledge, effectively mitigating mode collapse, and enhancing the diversity and informativeness of the generated sentences. 

  While this method exhibits significant improvements in the performance, it does require a certain amount of target domain images to learn domain-specific knowledge. 
  As a result, it is better suited for deployment in large-scale image captioning scenarios. 
  Nonetheless, the efficiency of GeneIC's parameters allows for straightforward future extensions. 
  For example, there is room for exploring further improvements in data utilization, and without retraining all parameters. 
  In conclusion, we hope that the empirical findings presented in this paper will make a valuable contribution to the advancement of general domain image captioning.

\appendix

\section*{Appendix}
  \section{Details of Multi-modal Large Language Models} \label{appx:more_details}
  In this paper, GeneIC employs BLIP2\_2.7B as the backbone, with OPT2.7B~\citep{abs-2205-01068} serving as the language decoder. 
  ZeroCap adopts GPT-2~\citep{radford2019language} as its language decoder, while Flamingo\_9B utilizes LLaMa-7B~\citep{abs-2302-13971} for the same purpose. 
  Moreover, BLIP2\_2.7B and BLIP2\_6.7B both rely on OPT2.7B and OPT6.7B as their respective language decoders. 
  As for MiniGPT4\_7B and InstructBLIP\_7B, they use Vicuna7b~\citep{vicuna2023} as their language decoder. 

  \section{Comparisons with Other Hand-crafted Prompts} \label{appx:prompt}
  Table~\ref{tab:prompt} provides a summary of the results obtained by the baseline model BLIP2\_2.7B using different hand-crafted prompts. 
  The findings reveal the model's sensitivity to hand-crafted prompts. 
  For example, when evaluating the CIDEr metric on the CUB-200 dataset, ``A bird photo depict'' exhibited a decrease of 3.4 in comparison to ``A photo of''. 
  However, for the CLIP-S metric, it showed an increase of 1.7. 
  The difference arises because ``A bird photo depict'' incorporates prior knowledge about the dataset, facilitating a more comprehensive description, but it may generate sentences with styles different from the ground truth. 
  In the Oxford-102 dataset, this phenomenon doesn't happen because it usually has close-up shots of flowers with less background information. 
  Randomly prompt vectors exhibit the poorest performance across most metrics, as expected, owing to their lack of prior knowledge. 
  In contrast, GeneIC demonstrates superior performance on most metrics, particularly on supervised metrics. 
  This indicates that through the exploration of variable and invariant features in target domain images, learned prompt vectors can effectively acquire domain-specific knowledge and guide the model towards improved generalization in the target domain. 
  Although GeneIC slightly lags behind ``A bird photo depict'' on the CLIP-S metric of the CUB-200 dataset, it still outperforms other hand-crafted prompts on the CLIP-S metric. 
  This indicates that the semantic consistency constraint ensures the model's acquisition of domain-specific knowledge without disregarding the background information in the images. 

  \section{Domain-specific Prompt Vectors vs Unified Prompt Vectors} \label{appx:domain_specific_prompt}
  Table~\ref{tab:change_vec} demonstrates the efficacy of domain-specific prompt vectors. 
  When applying the prompt vectors trained on the CUB-200 target domain to other target domains, the results underperform compared to utilizing prompt vectors specifically trained on the respective target domains. 
  This phenomenon indicates that knowledge from different target domains is not universally transferable. 
  
  \section{Visualization of Comparison Methods and Attribute-transferred Images} \label{appx:more_vis}
  Figure~\ref{fig:case_app} showcases examples generated by GeneIC and the comparison methods ($i.e.$, Multi-task, MiniGPT4\_7B, and InstructBLIP\_7B). 
  Compared to cross-domain methods ($i.e.$, Multi-task), the sentences generated by GeneIC exhibit more informative content and demonstrate the ability to generalize to any target domain, even in the absence of annotated data. 
  MiniGPT4\_7B, InstructBLIP\_7B, and GeneIC employ similar backbone networks, $i.e.$, pre-trained visual encoders + QFormer + pre-trained large language models. 
  However, MiniGPT4\_7B and InstructBLIP\_7B enhance the model's generation capability and its robust to hand-crafted prompts through fine-tuning the model with an instruction dataset , which consist of images and instructions.
  As observed from the Figure~\ref{fig:case_app}, across all datasets, the fine-grained information of sentences generated by MiniGPT4\_7B and InstructBLIP\_7B is superior to that of cross-domain methods. 
  However, compared to GeneIC, MiniGPT4\_7B and InstructBLIP\_7B exhibit hallucinations, generating sentences that include content not present in the images. 
  Meanwhile, it is essential to note that constructing instruction datasets comes with a substantial cost. 
  In contrast, GeneIC achieves an enhanced generalization performance for image captioning on target domain data almost at no cost. 
  Additionally, during the training process, GeneIC freezes most of the model parameters and only optimizes the prompt vectors, reducing training costs.

  Figure~\ref{fig:case_attr} displays more examples of attribute-transferred images. 
  It becomes evident that the main object in the attribute-transferred images undergoes significant attribute change in comparison to the original images, such as the color of birds and flowers. 
  GeneIC investigates both variable and invariant features within the target domain by analyzing the changes between the original images and attribute-transferred images. 
  By employing attribute consistency, it fine-tunes the prompt vectors to acquire domain-specific knowledge. 

\bibliography{ms}
\end{document}